\documentclass[acmtog, nonacm=true]{acmart}

\usepackage{booktabs} 
\usepackage[ruled]{algorithm2e} 

\SetAlFnt{\small}
\SetAlCapFnt{\small}
\SetAlCapNameFnt{\small}
\SetAlCapHSkip{0pt}
\usepackage{multirow}
\usepackage{relsize}
\usepackage{colortbl}

\newcommand{\hl}[1]{{#1}}

\begin{document}
\title{EgoLocate: Real-time Motion Capture, Localization, and Mapping with Sparse Body-mounted Sensors}

\author{Xinyu Yi}
\affiliation{%
  \institution{School of Software and BNRist, Tsinghua University}
  \country{China}}
\email{yixy20@mails.tsinghua.edu.cn}

\author{Yuxiao Zhou}
\affiliation{%
  \institution{ETH Zurich}
  \country{Switzerland}}

\author{Marc Habermann}
\affiliation{%
  \institution{Max Planck Institute for Informatics, Saarland Informatics Campus}
  \country{Germany}}

\author{Vladislav Golyanik}
\affiliation{%
  \institution{Max Planck Institute for Informatics, Saarland Informatics Campus}
  \country{Germany}}

\author{Shaohua Pan}
\affiliation{%
  \institution{School of Software and BNRist, Tsinghua University}
  \country{China}}

\author{Christian Theobalt}
\affiliation{%
  \institution{Max Planck Institute for Informatics, Saarland Informatics Campus}
  \country{Germany}}

\author{Feng Xu}
\affiliation{%
  \institution{School of Software and BNRist, Tsinghua University}
  \country{China}}
\email{xufeng2003@gmail.com}

\begin{abstract}
    Human and environment sensing are two important topics in Computer Vision and Graphics.
Human motion is often captured by inertial sensors, while the environment is mostly reconstructed using cameras.
We integrate the two techniques together in EgoLocate, a system that simultaneously performs human motion capture (mocap), localization, and mapping in real time from sparse body-mounted sensors, including 6 inertial measurement units (IMUs) and a monocular phone camera. 
On one hand, inertial mocap suffers from large translation drift due to the lack of the global positioning signal. 
EgoLocate leverages image-based simultaneous localization and mapping (SLAM) techniques to locate the human in the reconstructed scene.
On the other hand, SLAM often fails when the visual feature is poor. 
EgoLocate involves inertial mocap to provide a strong prior for the camera motion.
Experiments show that localization, a key challenge for both two fields, is largely improved by our technique, compared with the state of the art of the two fields.
%
Our codes are available for research at \url{https://xinyu-yi.github.io/EgoLocate/}.
\end{abstract}

\begin{teaserfigure}
  \includegraphics[width=\textwidth]{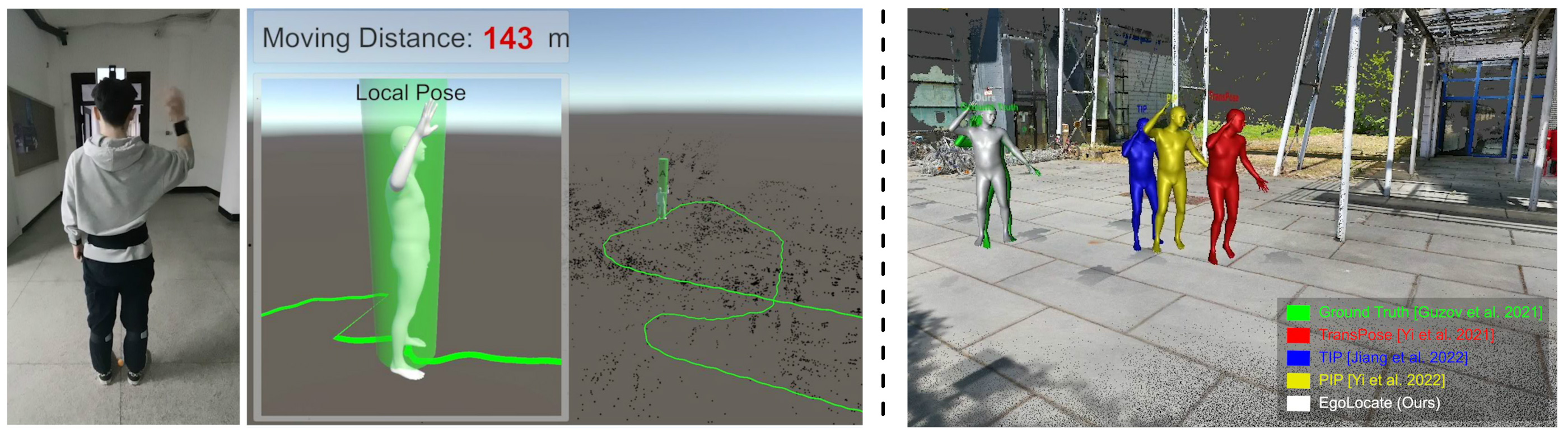}
  \caption{Key results of our method. \textit{Left}: for a user moving in a 3D scene, our technique estimates the user's pose, localizes the user's positions in the scene (green moving trajectory, loop closed after 143m motion with upstairs and downstairs), and reconstructs sparse 3D points of the scene (black scene points) simultaneously in real time. \textit{Right}: benefiting from the sensor fusion, the localization accuracy of our method is largely improved compared with the state-of-the-art techniques.}
  \label{fig:teaser}
\end{teaserfigure}

\maketitle
%
%
\section{Introduction}
%
%
%
%
Humans and environments are two important components of our world. 
Human motion sensing, aiming to capture and understand human activities, is a widely studied topic with many applications such as VR/AR, sports, and gaming.
Meanwhile, environment sensing techniques, which get the knowledge of the surrounding environment including its 3D shape and semantics, have long been studied and are widely used in areas such as autonomous driving and robot navigation. 
\par
While humans and environments are two indispensable and interdependent parts of the world, existing techniques mostly handle them independently. 
We believe that the combinative sensing of human motion and environments is of great importance for scenarios where humans interact with the environment.
And it may benefit new applications such as human behavior understanding, human localization, and motion planning.
%
%
%
%
\par
As an important human sensing technique, human motion capture (mocap) aims to reconstruct kinematic body motions and is widely studied in the literature.
%
%
%
The works that leverage multi-view external cameras (some even with multiple markers on the human body) usually achieve robust and accurate results~\cite{Chen2020,Zhang2020,Dong2019,shao2022diffustereo,reddy2021tessetrack}.
However, the capture space is constrained in the camera frustums as the performers need to be visible to the cameras.
While egocentric works do not constrain the recording space~\cite{mo2cap2, tome2019xr, hakada2022unrealego} and achieve lightweight reconstruction and real-time performance~\cite{TransPose,TIP,PIP}, they often suffer from translation drifts as there are no global positioning signals, which results in the wrong position of the human after a period of motion.
%
%
%
%
%
%
%
%
\par
For environment sensing, simultaneous localization and mapping (SLAM) is the capstone technique that leverages a moving camera to reconstruct the 3D scene and locate the camera in the scene simultaneously. 
In this setting, camera localization is a key as it is required to fuse the 3D information at different time instances together to make a complete 3D scene.
Despite previous efforts, current pure visual SLAM~\cite{monoslam,PTAM,lsdslam,svo,orbslam2} may still fail to track the camera when feature absence or motion blur happens. 
%
%
By incorporating an IMU into the camera, visual-inertial SLAM becomes more robust when reliable visual features are insufficient~\cite{msckf,okvis,vinsmono,ORBSLAM3}, but still suffers from tracking losses in long-time camera occlusion and fast camera rotation, especially when the frequency of the IMU signals is low.
This will also result in large localization errors or even system failure.
%
%
%
%
%
%
%
%
%
%
%
\par
Localization is a key task for both mocap and SLAM and is very challenging.
While inertial-sensor-based mocap explores \textit{inner information} such as human motion signals and motion priors, SLAM majorly relies on \textit{outer information}, \textit{i.e.}, environment captured by the camera. 
The former achieves good stability, but the global position drifts accumulate in long-time motions as no correction from outside is available; the latter estimates the global position in the scene with high accuracy, but suffers from tracking losses when the environmental information is unreliable, \textit{e.g.}, in textureless or occluded places.
%
%
To this end, we propose EgoLocate, which effectively combines the two complementary techniques (mocap and SLAM) together and achieves robust and accurate localization (see Fig.~\ref{fig:teaser}). 
%
%
Our system leverages sparse body-mounted sensors including 6 IMUs mounted on the forearms, the lower legs, the head, and the pelvis of the human, and a monocular RGB camera attached to the head looking outwards.
This design is inspired by the behavior of real humans: when humans are in a new environment, they use their eyes to get knowledge of the scene with their location and plan their movements in the scene.
The monocular camera acts as the "eye" that provides visual signals to our method for real-time scene reconstruction and self-localization, while the IMUs measure human motion in the scene. 
\par
We propose an organically coupled framework to exploit the complementary advantages of the sparse inertial mocap and SLAM techniques.
In this framework, the human motion prior is integrated into multiple key components of SLAM, and the SLAM localization is fed back to the mocap routine.
By jointly optimizing the camera pose and map point positions with the awareness of mocap, both the tracking and mapping are improved in terms of accuracy and robustness.
When the visual cues are reliable, the SLAM is able to correct the drift of the mocap using the environmental information;
when the visual features are sparse due to camera occlusion or extreme lighting, the mocap module can provide pose and translation estimation for the SLAM system, avoiding total failure in previous SLAM systems.
%
%
Furthermore, we propose a mocap-related map point confidence, which helps to dynamically determine the importance of each map point in the bundle adjustment.
By reducing the influence of potentially wrong 3D points, our method achieves a good balance between the map point-based constraints and the mocap constraints, which helps to reduce the uncertainty in the tracking and improve the localization accuracy.
Combining the mocap and the SLAM system results in a win-win.
On the one hand, with the help of camera localization, the translation drift in inertial mocap is largely reduced.
On the other hand, with the help of mocap, the accuracy and robustness of localization and mapping are greatly improved.
\par
To the best of our knowledge, our method is the first effort that achieves real-time mocap and simultaneous mapping from sparse body-mounted sensors.
The most related works to ours are HPS~\cite{HPS} and HSC4D~\cite{HSC4D}.
HPS performs offline human motion capture and localization in a known scene from 17 densely placed IMUs and a body-mounted camera, and requires a pre-scanned scene with plenty of registered photos.
HSC4D leverages 17 IMUs and a LiDAR sensor placed on the human body, and estimates the human motion and the scene simultaneously but also in an offline manner.
Compared with these works, our system is much more \textit{lightweight}, as we only use 6 IMUs and a body-mounted monocular camera.
Importantly, we also do \textit{not} rely on pre-scanning the scene.
Moreover, our system performs motion capture, localization, and mapping in \textit{real time}.
\par
In summary, our contributions are:
\begin{itemize}
    \item We propose the first real-time simultaneous human motion capture, localization, and mapping system using only 6 IMUs and a body-mount camera, which estimates drift-free human motion and sparse scene points in unconstrained 3D space.
    \item We propose a tightly-coupled optimization framework by incorporating the mocap prior in key SLAM modules, which improves both the localization and mapping accuracy as well as the robustness \hl{to poor visual features}.
    \item We propose mocap-related map point confidence to dynamically adjust the influence of each map point and the mocap constraints in the bundle adjustment, which improves the localization accuracy while reducing the uncertainty in the results.
\end{itemize}
\par
Our work is trying to merge human sensing and environment sensing together.
%
%
Even though we mostly focus on localization, we believe this work is a first step toward joint motion capture and dense environment sensing and reconstruction.
\section{Related Work}
\subsection{Egocentric Human Motion Capture}
Egocentric human motion capture (mocap) is widely studied due to its advantage that the recording space is not constrained in a fixed volume.
Existing works leverage different body-mount sensors to capture or infer human motions.
One category of works uses body-mounted cameras, either capturing the human motion from egocentric views \hl{(cameras looking at the wearer)}~\cite{EgoCap,SelfPose2020,Wang2021,hakada2022unrealego,tome2019xr,Hori2022,wang2023scene} or estimating the human motion from first-person views (cameras looking outwards)~\cite{Jiang2017Ego,Yuan2018,Yuan2019,Shiratori2011,li2022ego,Luo2021DynamicsRegulatedKP}.
\hl{These techniques enable space-free human mocap and have yielded many impressive results.
However, the former setting may encounter self-occlusions and difficulty in detecting global movements, limiting the accuracy of pose and translation estimation.
While the latter is limited in its ability to reconstruct certain types of human motion as the human body is not visible in the camera view.}
A combination of egocentric-view and first-person-view cameras can help with translation estimation~\cite{Liu2022ego}.
\hl{Some approaches~\cite{VIP,yuan2022glamr,BodySLAM,Liu20214D} leverage a moving external camera, \textit{e.g.}, held by another person, to capture the human motion. 
They do an excellent job of decoupling human and camera motion, \textit{e.g.,} by leveraging/reconstructing the static background scene.
Overall, these camera-based approaches have made remarkable progress in enabling space-free human motion capture.}
Another category of works captures human motions from sparse non-vision-based body-mounted sensors, including accelerometers~\cite{Slyper2008,Tautges2011,Riaz2015}, inertial measurement units (IMUs)~\cite{SIP,DIP,Geissinger2020,Patrik2021,TransPose,PIP,TIP,Vlasic2007,Xia2022,IEEEVRIMUego}, and electromagnetic sensors~\cite{empose}.
They overcome common limitations of vision-based approaches like occlusions, and can estimate accurate human articulated pose and global movements within a short period after calibration. 
However, since there are no global positioning signals from the sensors, their translation drifts can be large, resulting in the wrong position of the human after a long period of motion.
This can lead to penetration and even wrong interaction with the digital scene, which is undesirable in applications like VR and gaming.
Some works leverage body-mounted VR devices~\cite{avatarposer,questsim} to capture the human motion in a constrained space.
HPS~\cite{HPS} and its extension~\cite{HOPS} propose to locate the human in the environment by leveraging a head-mounted camera and densely-worn IMUs.
They achieve good accuracy on human localization, but they rely on the pre-scanned scene with plenty of registered photos and run in an offline manner.
HSC4D~\cite{HSC4D} simultaneously reconstructs the human motion and the scene with good consistency.
However, they need dense IMUs and a body-worn LiDAR sensor and also run in an offline manner.
While we use \textit{sparse} IMUs and a body-mounted camera to capture the space-free human motion in \textit{real time}, without the need to pre-scan the scene.
%
%
\subsection{Visual and Visual-Inertial SLAM}
Simultaneous localization and mapping (SLAM) has long been studied due to its importance in scene reconstruction and localization in unknown environments. 
Visual SLAM~\cite{monoslam,lsdslam,svo,dso,dsm,orbslam2,droidslam,tandem,d3vo,codeslam} leverages a monocular camera and estimates the camera movements while simultaneously reconstructing the scene in 3D points or structures.
However, they often fail to track the camera when the visual feature is poor, \textit{e.g.}, due to motion blur or camera occlusion.
They are also unaware of the true scale of the scene and the camera movement.
Visual-inertial SLAM incorporates an IMU to the camera, which becomes more robust to fast camera motions and poor visual features, and can also determine the scale factor~\cite{msckf,okvis,orbslamvi,vidso,vinsmono,ORBSLAM3}.
However, they still suffer from tracking failures when the camera is occluded for a longer time, and are also sensitive to \hl{the camera-IMU relative pose (IMU extrinsic) and IMU noise parameters (IMU intrinsic)}.
Some works~\cite{ORBSLAM3,castle2008video,schmuck2017multi} use a multi-map system to improve the robustness.
\cite{ORBSLAM3} starts a new map when the tracking is lost.
However, the online localization results will be reset in the new map, which means it is not possible to obtain the absolute position in real time until the new map is merged with the initial map.
Our system focuses on situations where humans wear the camera. 
In this situation, the existing SLAM techniques are mostly unaware of the human prior.
To this end, we propose to incorporate inertial human mocap with SLAM techniques, which not only improve the accuracy of the localization and mapping, but also enhance the robustness of the system.
%

\section{Method}
\begin{figure*}[t]
  \includegraphics[width=\linewidth]{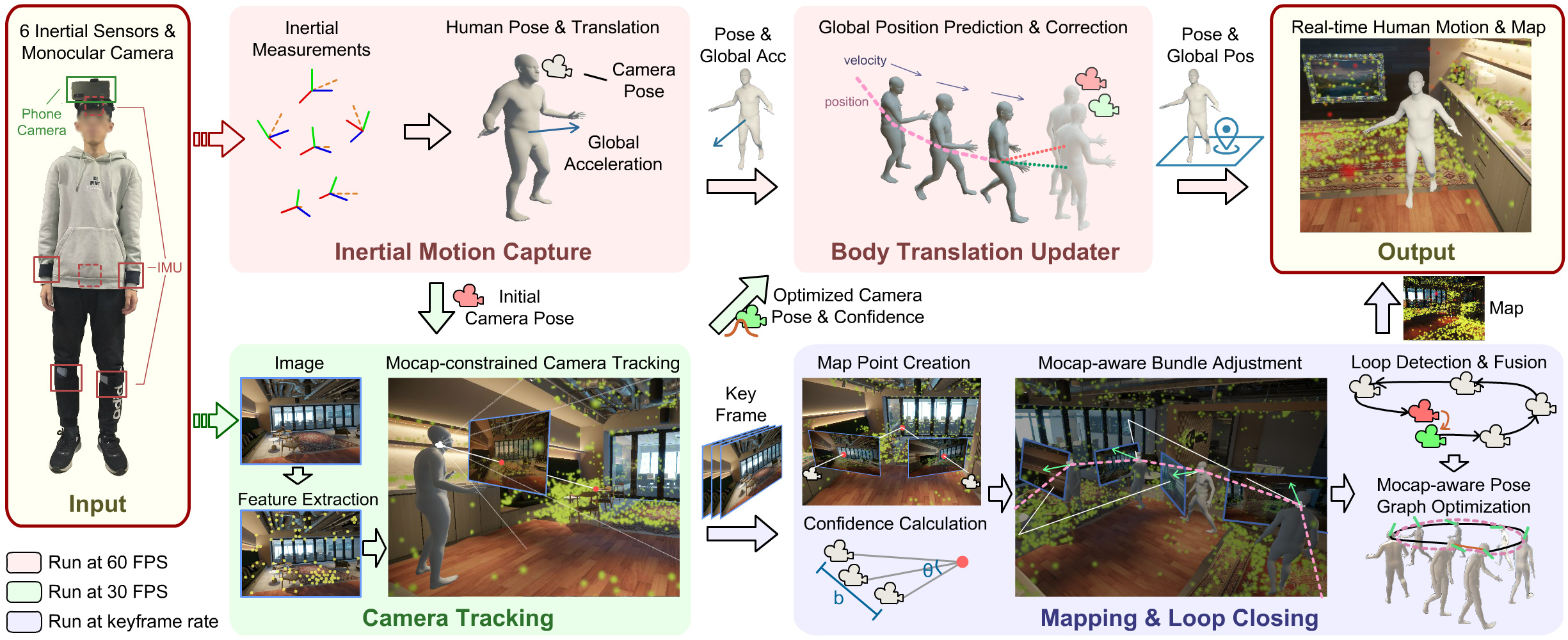}
        \caption{\textbf{Overview of our method.} The inputs are real-time measurements of 6 IMUs and images from a body-worn camera. We first estimate initial human poses and camera poses from sparse inertial signals (Inertial Motion Capture, Sec.~\ref{sec:method-motion-capture}). Next, we refine the camera pose by minimizing the mocap constraints and the reprojection errors of the 3D-2D feature matches (Camera Tracking, Sec.~\ref{sec:method-camera-tracking}). Then, the refined camera pose and a confidence are used to correct the human's global position and velocity (Body Translation Updater, Sec.~\ref{sec:method-translation-updater}). This results in 60-FPS drift-free human motion output. In parallel, we perform mapping and loop closing using keyframes (Mapping \& Loop Closing, Sec.~\ref{sec:method-mapping-and-loop-closing}). We calculate another confidence for each map point, and use it in bundle adjustment (BA), where we jointly optimize the map points and the camera poses by a combined mocap and weighted reprojection error. If a loop is detected, we also perform pose graph optimization with mocap constraints. The estimated map is also outputted in real time.}
  \label{fig:method}
\end{figure*}
Our goal is to use body-worn sensors to capture the body motion while simultaneously reconstructing the environment and locating the human in the environment in real time.
The input of our system is the synchronized signals from the sensors, including the inertial measurements from 6 IMUs and the monocular images from a head-mounted camera.
The system estimates the human pose and global motion, as well as the reconstructed scene as a sparse 3D point cloud; see Fig.~\ref{fig:method} for an overview. 
Our method first performs human motion capture from the inertial inputs (Sec.~\ref{sec:method-motion-capture}).
Then, the camera pose is optimized by minimizing the reprojection error of the reconstructed map with mocap constraints (Sec.~\ref{sec:method-camera-tracking}).
The optimized camera pose is then used to update the human translation to avoid the drifting artifacts caused by inertial error accumulation (Sec.~\ref{sec:method-translation-updater}).
In parallel to the motion tracking, our method reconstructs the scene map and performs loop closing with mocap awareness (Sec.~\ref{sec:method-mapping-and-loop-closing}).
The calibration of the system is discussed in Sec.~\ref{sec:method-initialization}, including how to initialize the inertial mocap and SLAM respectively, as well as how to calculate the extrinsic between the mocap and the SLAM system. 

\subsection{Inertial Motion Capture}\label{sec:method-motion-capture}
From the \hl{orientation and acceleration measurements} of 6 body-mounted IMUs, we first estimate an initial human pose and translation.
Our method follows the sparse inertial mocap work PIP~\cite{PIP}, but we remove its planar assumption that the whole scene is a known level ground and the human movements are on the ground because our goal is to enable unconstrained mocap in free 3D space.
We also re-design the motion optimizer in PIP by removing the force calculation.
This is because we do not know the dense geometry of the scene (the reconstruction is only in sparse scene points), making it impossible to detect human-environment collisions to apply forces on the contacts.
%
%
More details about the modification to PIP are presented in App.~\ref{app:method-motion-capture}.
\par
Our inertial mocap estimates human pose (in terms of joint angles) and the human global rotation and translation in a user-defined mocap coordinate system at 60 FPS.
Then, the 6-DoF pose of the head-mounted camera can be extracted from the inertial mocap result with the calibrated camera-head extrinsic (see Sec.~\ref{sec:method-initialization}), which will be used as an initialization of the following camera tracking step.
%
%
Note that the inertial mocap provides good articulated pose estimation, but the global motion of the human drifts significantly over time.
%
%

\subsection{Camera Tracking}\label{sec:method-camera-tracking}
%
%
This module aims to estimate better camera poses by using the 30-FPS images from the head-mounted camera and the estimated camera poses from the inertial mocap, which are synchronized to the same frame rate of images.
We design this module based on ORB-SLAM3~\cite{ORBSLAM3}, where we first extract ORB feature points (keypoints) from the image and find the matches between the keypoints and the 3D map points using feature similarity. 
Note that the map points are reconstructed and maintained online by a mapping module (see Sec.~\ref{sec:method-mapping-and-loop-closing}).
Then, we optimize the camera pose by minimizing the reprojection error of the matches with additional mocap constraints.
\subsubsection{Mocap-constrained Camera Tracking}
Formally, we denote the world position of the $i$th 3D map point as $\boldsymbol{x}_i^{\mathrm{3D}} \in \mathbb{R}^3$ and the pixel coordinates of the matched 2D keypoint as $\boldsymbol{x}_i^{\mathrm{2D}} \in \mathbb{R}^2$, where $i\in\mathcal{X}$ for all matches.
We use $\bar{\boldsymbol{R}} \in \mathrm{SO(3)}$ and $\bar{\boldsymbol{t}} \in \mathbb{R}^3$ to denote the camera pose \textit{w.r.t} the world before the optimization, which is obtained from the inertial mocap. 
The mocap-constrained camera tracking is then formulated as the nonlinear optimization of the camera pose $\boldsymbol{R},\boldsymbol{t}$:
\begin{equation}\label{eq:tracking}
\begin{array}{c}
    \mathop{\arg\min}\limits_{\boldsymbol{R},\boldsymbol{t}}\mathcal{E}_\mathrm{proj}(\boldsymbol{R},\boldsymbol{t}) + \mathcal{E}_\mathrm{mocap}(\boldsymbol{R},\boldsymbol{t})\\
    \mathcal{E}_\mathrm{proj} = \mathlarger{\sum}\limits_{i\in\mathcal{X}} \uprho\left(\left\|\boldsymbol{x}^{\mathrm{2D}}_i - \uppi\left(\boldsymbol{R}^T (\boldsymbol{x}^{\mathrm{3D}}_i - \boldsymbol{t})\right)\right\|^2_{\boldsymbol{\varSigma}_i}\right)\\
    \mathcal{E}_\mathrm{mocap}=\lambda_\mathrm{R}\left\|\mathrm{Log}(\bar{\boldsymbol{R}}^T\boldsymbol{R})\right\|^2 + \lambda_\mathrm{t}\left\|\bar{\boldsymbol{t}} - \boldsymbol{t}\right\|^2,
\end{array}
\end{equation}
where $\uprho$ is the Huber robust cost function, $\boldsymbol{\varSigma}_i$ is the covariance
matrix associated with the keypoint scale\footnote{The \textit{scale} of a keypoint is determined by the level where it is detected in the image pyramid. See~\cite{ORBSLAM}.}, $\mathrm{Log}:\mathrm{SO(3)}\rightarrow\mathbb{R}^3$ is a mapping from the Lie group to the vector space, and $\uppi$ is the camera projection function.
$\lambda_\mathrm{R}$ and $\lambda_\mathrm{t}$ are the coefficients aligning the unit and controlling the weights of the corresponding terms. 
The reprojection term $\mathcal{E}_\mathrm{proj}(\boldsymbol{R},\boldsymbol{t})$ refines the camera pose by minimizing the projection error of the matched 3D-2D points, while the mocap term $\mathcal{E}_\mathrm{mocap}(\boldsymbol{R},\boldsymbol{t})$ constrains the camera pose to be near to the mocap-based pose.
Similar to~\cite{ORBSLAM3}, we perform the optimization 3 times and classify the keypoint matches as inliers/outliers after each optimization, where outliers are excluded from the next optimization.
To this end, the mocap term provides a strong prior on the camera pose, which helps with the classification of the wrong 3D-2D matches and reduces the errors caused by outliers.
%
%
The coefficients of the mocap terms are set to $\lambda_\mathrm{R}=0.01f^2$ and $\lambda_\mathrm{t}=0.5f^2s^2$ where $f$ is the camera focal length and $s$ is the scale factor of the SLAM coordinate system computed in the calibration module (Sec.~\ref{sec:method-initialization}).
This automatically converts the unit of the mocap errors to pixels, which is at the same magnitude as the reprojection error.
Such a strategy makes the optimization independent with the camera intrinsic and the scale factor.
We leverage the Levenberg–Marquardt algorithm implemented in g2o~\cite{g2o} to solve the nonlinear optimization.
To ensure real-time performance, we calculate the Jacobian analytically, which is presented in App.~\ref{app:nonlinear-optimizations}, \hl{and retrieve a local visible map (instead of using the entire map) for the keypoint matching and camera tracking, which is detailed in App.~\ref{app:local-map}.} 
After the optimization, we extract the number of the inlier 3D-2D matches $n$ as the confidence to indicate the quality of the refined camera pose, which will be used in the following module. 
Intuitively, we have larger confidence if more map points have been successfully matched.
%
\subsubsection{Camera Pose Alignment}
\hl{
In practice, the absolute camera translation derived from pure inertial mocap usually contains large drifts due to inertial error accumulation. 
As a result, directly using it in the camera tracking optimization is not feasible for long sequences because such non-linear optimizations typically require a good initialization.
Therefore, before optimization, we perform a camera pose alignment step for each frame to reduce the drifts
%
Specifically, we compute the \textit{relative} camera rotation and translation from the mocap and add it to the previous SLAM-optimized camera pose. 
This is based on the observation that the refined camera poses in SLAM consider the visual information and the drift problem is largely reduced.
%
}
%
%
%
%
%
%
We denote the SLAM-optimized camera pose as $\boldsymbol{R} \in \mathrm{SO(3)}, \boldsymbol{t} \in \mathbb{R}^3$ for the camera orientation and position \textit{w.r.t} the world respectively, and the camera pose extracted from mocap as $\tilde{\boldsymbol{R}}, \tilde{\boldsymbol{t}}$.
Then, the alignment can be written as:
\begin{equation}
\begin{array}{cc}
    \bar{\boldsymbol{R}}_{\mathrm{cur}} = \boldsymbol{R}_{\mathrm{last}} \tilde{\boldsymbol{R}}_{\mathrm{last}}^T \tilde{\boldsymbol{R}}_{\mathrm{cur}}\\
    \bar{\boldsymbol{t}}_{\mathrm{cur}} = \boldsymbol{t}_{\mathrm{last}} - \tilde{\boldsymbol{t}}_{\mathrm{last}} + \tilde{\boldsymbol{t}}_{\mathrm{cur}},
\end{array}
\end{equation}
where $\bar{\boldsymbol{R}}_{\mathrm{cur}}$ and $\bar{\boldsymbol{t}}_{\mathrm{cur}}$ are the aligned camera pose at the current frame, subscript $\cdot_{\mathrm{cur}}$ and $\cdot_{\mathrm{last}}$ denote the current and last frame respectively.
We use the last \textit{keyframe} (defined later) for the alignment because its pose is more accurate as it has been optimized in the bundle adjustment (BA) (Sec.~\ref{sec:method-mapping-and-loop-closing}).
However, if a relocalization\footnote{\hl{\textit{Relocalization} refers to the process of re-estimating the camera pose in the built map when the visual tracking is lost. See~\cite{ORBSLAM}.}} recently happens, when no keyframe is available after the relocalization, we choose to use the latest \textit{frame} after the relocalization to avoid the drift.
As the orientation estimation $\tilde{\boldsymbol{R}}_{\mathrm{cur}}$ does not suffer from evident drifts in inertial mocap, we finally always rotate $\bar{\boldsymbol{R}}_{\mathrm{cur}}$ towards $\tilde{\boldsymbol{R}}_{\mathrm{cur}}$ by $0.1$ through spherical linear interpolation.
\hl{This ensures that the aligned orientation does not deviate from the mocap estimation significantly.}
After the alignment, the drifts in the camera pose are mostly removed, and then we can perform the keypoint matching and the camera tracking (Eq.~\ref{eq:tracking}).
\subsubsection{Keyframe Decision}
This module also determines whether the current frame becomes a \textit{keyframe}.
Keyframes are a set of representative frames with minimal redundancy on visual and mocap information, and are mainly used for mapping and loop closing.
This is a key to achieving real-time performance~\cite{PTAM}.
We select the keyframes following~\cite{ORBSLAM3}, \hl{where multiple criteria are considered, such as sufficient time interval between keyframes, sufficient matched keypoints, and enough visual changes}.
We also include the mocap-derived camera poses $\tilde{\boldsymbol{R}}, \tilde{\boldsymbol{t}}$ in the keyframes, as they provide mocap priors for the global optimizations detailed later.

\subsection{Mapping and Loop Closing}\label{sec:method-mapping-and-loop-closing}
In this section, we involve mocap information in the global optimization schemes in traditional SLAM, including the bundle adjustment (BA) scheme and the pose graph optimization in loop closure, which leads to more accurate mapping and localization results.
%
%
%
%
Specifically, we integrate the mocap prior in the BA algorithm, where the map and keyframe poses\footnote{\hl{For brevity, we use \textit{keyframe pose} to denote the camera pose (position and orientation) at the keyframe throughout the paper.}} are optimized by considering both 3D reprojection errors and mocap constraints (Sec.~\ref{sec:method-mapping-and-loop-closing-2}).
In BA, we calculate mocap-related map point confidences to dynamically match the weight of the projection error of each map point with the mocap constraints (Sec.~\ref{sec:method-mapping-and-loop-closing-1}).
%
%
Besides, when a loop closure is detected (\textit{i.e.}, the camera returns to a visited place), we perform a \textit{pose graph optimization} to close the loop, where the mocap prior is also incorporated, which constrains the relative pose between each keyframe from the view of mocap (Sec.~\ref{sec:method-mapping-and-loop-closing-3}).
%
%
%
%
%
%
\subsubsection{Map Point Confidence Calculation}\label{sec:method-mapping-and-loop-closing-1}
\begin{figure}[t]
  \includegraphics[width=\linewidth]{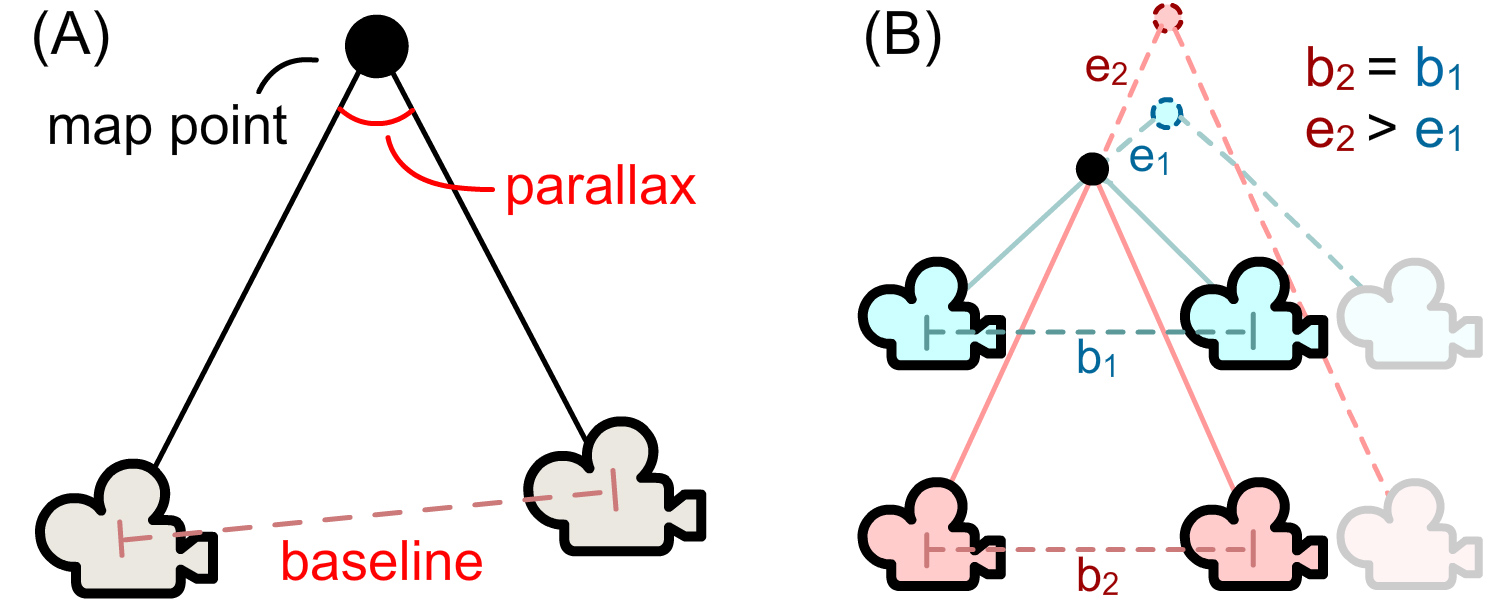}
        \caption{Explanation on the mocap-related map point confidence. (A) The confidence is computed from the baseline length and the parallax of the keyframes that observe the point. (B) With the same baseline length $b_1=b_2$ and the same perturbation on the camera positions (denoted as the transparent cameras), a larger parallax (blue cameras) is often less prone to errors compared with a smaller parallax (red cameras) (\textit{i.e.}, point error $e_1<e_2$).}
  \label{fig:parallex}
\end{figure}
%
In order to achieve accurate results in BA, it is essential to consider the relative weights of mocap constraints in comparison to traditional reprojection terms. 
To make a delicate design of the weights, we no longer treat all the map points equally but calculate their individual confidence to match their relative weights to the mocap constraints.
%
%
We consider a 3D point as accurate if \textit{1)} the frames that observe it are located in largely different positions in the real world and \textit{2)} the viewing direction in which the frames look at the point span a sufficiently large angle.
This is inspired by the multi-view 3D reconstruction, where a larger baseline and parallax lead to accurate estimation (see Fig.~\ref{fig:parallex}).
%
%
%
%
\par
%
%
%
%
%
    %
%
%
To be specific, \hl{for each map point $i$ that will participate in the BA optimization}, we first calculate the baseline length $b_i$ in the mocap coordinate system and the parallax angle $\theta_i$ of multiple keyframes in which the point $i$ is visible.
Then, the map point confidence $c_i$ is calculated as the scaled product of the baseline length and the parallax:
\begin{equation}
    c_i=kb_i\theta_i,
\end{equation}
where $k$ is experimentally set to 50 to make the confidence near 1 for most of the points.
The transformation to the mocap coordinate system is necessary because the confidence should be invariant to the scale factor of SLAM.
This is a requirement to combine mocap and SLAM in BA.
%
%
\subsubsection{Mocap-aware Bundle Adjustment}\label{sec:method-mapping-and-loop-closing-2}
\begin{figure}[t]
  \includegraphics[width=0.94\linewidth]{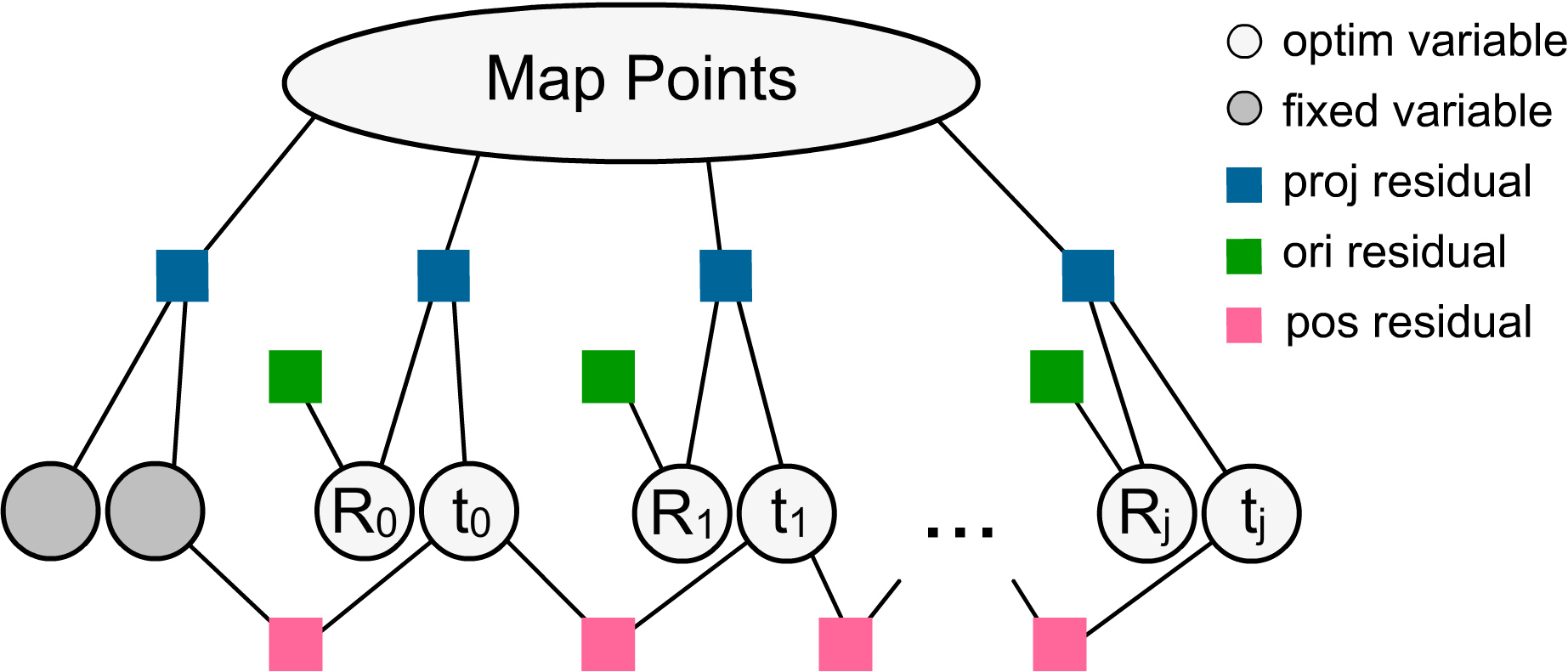}
        \caption{Factor graph for the mocap-aware bundle adjustment. Circles denote optimizable/fixed variables (\textit{i.e.}, map points and keyframe poses). Squares denote constraints among the connected variables (\textit{i.e.}, reprojection residual in blue for the observation constraints, mocap residual in green for the orientation constraints and pink for the position constraints).}
  \label{fig:factor}
\end{figure}
With the map point confidence, we perform a mocap-aware BA to refine the recent keyframe poses and the visible map.
The variables to be optimized include the poses of the recent keyframes and the positions of the map points observed by these keyframes.
Other keyframes that also observe the points also participate in the BA, but remain fixed.
We denote the optimizable and fixed set of keyframe poses as $\mathcal{K}_\mathrm{o}$ and $\mathcal{K}_\mathrm{f}$ respectively, and the observed map points of keyframe $j$ as $\mathcal{X}_j$.
Let $\mathcal{R}_\mathrm{o}=\{\boldsymbol{R}_j | j \in \mathcal{K}_\mathrm{o}\}$ and $\mathcal{T}_\mathrm{o}=\{\boldsymbol{t}_j | j \in \mathcal{K}_\mathrm{o}\}$ denote the optimizable keyframe poses \textit{w.r.t} the world, and $\mathcal{X}^{\mathrm{3D}}=\bigcup_{j \in \mathcal{K}_\mathrm{o}}\{\boldsymbol{x}_i^{\mathrm{3D}}|i\in \mathcal{X}_j\}$ denotes the map point positions in the world.
The mocap-aware BA is then defined as:
\begin{equation} \label{eq:ba}
\begin{array}{c}
    \mathop{\arg\min}\limits_{\mathcal{R}_\mathrm{o},\mathcal{T}_\mathrm{o},\mathcal{X}^{\mathrm{3D}}}\mathcal{E}_\mathrm{proj}(\mathcal{R}_\mathrm{o},\mathcal{T}_\mathrm{o},\mathcal{X}^{\mathrm{3D}}) + \mathcal{E}_\mathrm{mocap}^\mathrm{(R)}(\mathcal{R}_\mathrm{o}) + \mathcal{E}_\mathrm{mocap}^\mathrm{(t)}(\mathcal{T}_\mathrm{o})\\
    \mathcal{E}_\mathrm{proj} = \mathlarger{\sum}\limits_{j\in\mathcal{K}_\mathrm{o}\cup\mathcal{K}_\mathrm{f}}\mathlarger{\sum}\limits_{i\in\mathcal{X}_j} \uprho\left(c_i\left\|\boldsymbol{x}^{\mathrm{2D}}_i - \uppi\left(\boldsymbol{R}_j^T (\boldsymbol{x}^{\mathrm{3D}}_i - \boldsymbol{t}_j)\right)\right\|^2_{\boldsymbol{\varSigma}_i}\right)\\
    \mathcal{E}_\mathrm{mocap} ^\mathrm{(R)} = \mu_\mathrm{R}\mathlarger{\sum}\limits_{j\in\mathcal{K}_\mathrm{o}}\left\|\mathrm{Log}(\tilde{\boldsymbol{R}}_j^T\boldsymbol{R}_j)\right\|^2 \\
    \mathcal{E}_\mathrm{mocap} ^\mathrm{(t)} = \mu_\mathrm{t}\mathlarger{\sum}\limits_{j\in\mathcal{K}_\mathrm{o}}\left\|(\tilde{\boldsymbol{t}}_j-\tilde{\boldsymbol{t}}_{\mathrm{prev}(j)}) - (\boldsymbol{t}_j-\boldsymbol{t}_{\mathrm{prev}(j)})\right\|^2,
\end{array}
\end{equation}
where $\mathrm{prev}(j)$ denotes the previous keyframe of keyframe $j$, $\tilde{\boldsymbol{R}}$, $\tilde{\boldsymbol{t}}$ are the initial camera pose obtained from the mocap, and the coefficients of the mocap terms are set to $\mu_\mathrm{R}=0.01f^2$ and $\mu_\mathrm{t}=0.05f^2s^2$ to align the unit to pixels and make them scale-invariant.
The mocap-aware BA optimization is easier to understand through the factor graph as shown in Fig.~\ref{fig:factor}.
The reprojection term $\mathcal{E}_\mathrm{proj}$ in Eq.~\ref{eq:ba} corresponds to the blue nodes in the factor graph, which computes the reprojection error using the keyframe poses and the map points (shown as the adjacent variables connected to the node).
The mocap terms involve a rotation term $\mathcal{E}_\mathrm{mocap} ^\mathrm{(R)}$ (green nodes) that constrains the keyframe rotations to be near to the mocap results, and a translation term $\mathcal{E}_\mathrm{mocap} ^\mathrm{(t)}$ (pink nodes) that constrains the translations between adjacent keyframes to be similar to the mocap estimation. 
We apply \textit{absolute} constraints on the keyframe rotations but \textit{relative} constraints on the keyframe translations because the translation estimation suffers from drifts over time while the rotation estimation is relatively stable in the inertial mocap.
\par
Note that the map point confidence $c_i$ is used as the weights of the reprojection error, where potentially good points have larger weights compared with the mocap priors, which help to refine the keyframe poses, while probably inaccurate points (\textit{e.g.}, recently triangulated points) have smaller weights compared with the mocap priors, which can be refined in the optimization and have a lower impact in estimating the keyframe poses.
With the help of the mocap priors and the map point confidence, BA becomes more stable, and both the mapping and keyframe pose accuracy can be improved.
\par
We solve the nonlinear problem using graph optimization implemented in g2o~\cite{g2o}.
To accelerate the algorithm, the map points are marginalized as in~\cite{ORBSLAM3}, and we calculate the Jacobian of the keyframe poses and map points analytically, which is presented in App.~\ref{app:nonlinear-optimizations}.
As our system targets at \textit{online} localization, the optimized keyframe poses after the BA are not directly used in the real-time tracking system due to the delay.
%
%
\hl{
Specifically, it affects the real-time tracking implicitly: the mapping module updates the keyframe camera poses and the map points in parallel with the real-time modules in another thread, and the real-time camera tracking module tracks the camera pose \textit{w.r.t} the \textit{updated} map.}
Thus, our real-time results can be improved. 

\subsubsection{Mocap-aware Pose Graph Optimization}\label{sec:method-mapping-and-loop-closing-3}
When a loop is detected, we perform a pose graph optimization to close the loop.
This optimization involves only keyframe poses but no map points, which aims at closing the loop by distributing the aligning error to all keyframes.
Based on ORB-SLAM3~\cite{ORBSLAM3}, we perform the optimization on the \textit{essential graph} over the keyframe poses expressed as rigid transformations in $\mathrm{SE}(3)$ (rather than similarity transformations in $\mathrm{Sim}(3)$) as we do not suffer from the scale drift due to the combination with mocap.
We denote the keyframes (vertices) and connections (edges) of the essential graph as $\mathcal{F}$ and $\mathcal{C}$ respectively.
%
%
The pose graph optimization is then defined as:
\begin{equation}
    \begin{array}{c}
    \mathop{\arg\min}\limits_{\{\boldsymbol{T}_j\}_{j\in\mathcal{F}}}\mathcal{E}_\mathrm{graph} + \mathcal{E}_\mathrm{mocap}\\
    \mathcal{E}_\mathrm{graph} = \mathlarger{\sum}\limits_{(i,j)\in\mathcal{C}}\left\|\mathrm{Log}(\boldsymbol{T}_{ij}\boldsymbol{T}_j^{-1}\boldsymbol{T}_i)\right\|^2\\
    \mathcal{E}_\mathrm{mocap} = \omega_\mathrm{pose}\mathlarger{\sum}\limits_{j\in\mathcal{F}}\left\|\mathrm{Log}(\tilde{\boldsymbol{T}}_{\mathrm{prev}(j)}^{-1}\tilde{\boldsymbol{T}}_j\boldsymbol{T}_j^{-1}\boldsymbol{T}_{\mathrm{prev}(j)})\right\|^2,
\end{array}
\end{equation}
where $\boldsymbol{T}_j \in \mathrm{SE}(3)$ is the pose of keyframe $j$ \textit{w.r.t} the world, $\boldsymbol{T}_{ij} \in \mathrm{SE}(3)$ is the relative transformation between keyframe $i$ and $j$ before the optimization,  $\tilde{\boldsymbol{T}}_j$ is the transformation obtained by mocap, $\mathrm{Log}:\mathrm{SE(3)}\rightarrow\mathbb{R}^6$ maps a rigid transformation to the vector space, and $\omega_{\mathrm{pose}}$ is the coefficient of the mocap term, which we empirically set to $0.2$.
Intuitively, the graph term $\mathcal{E}_\mathrm{graph}$ constrains the change in the relative pose between the connected keyframes to be small before and after the optimization, while the mocap term regularizes the keyframe 6-DoF trajectory to be similar to the estimation of the mocap in a local manner.
%
%

\subsection{Body Translation Updater}\label{sec:method-translation-updater}
The camera tracking module considers both visual and inertial information, so the output 30-FPS camera localization should be able to refine the 60-FPS human motion from the inertial motion capture module.
%
%
%
%
To make the refinement fully constrained, we only update the human translation and implement the module using a prediction-correction algorithm in Kalman Filter. 
\par
Specifically, we define the global position and velocity of the human as the state variables $(\boldsymbol{p},\boldsymbol{v})$, which will be refined in this module.
Using the global acceleration $\boldsymbol{a}$ extracted from the mocap module, the state variables can be predicted by the following prediction equation:
\begin{equation}\label{eq:kf}
\begin{array}{c}
    \boldsymbol{p}_k = \boldsymbol{p}_{k-1} + \boldsymbol{v}_{k-1}\Delta t \\
    \boldsymbol{v}_k = \boldsymbol{v}_{k-1} + \boldsymbol{a}_{k-1}\Delta t + \boldsymbol{\varphi}_{k-1}, \\
\end{array}
\end{equation}
where the subscript $k$ indicates the $k$th frame, $\Delta t=1/60$ is the time interval, and $\boldsymbol{\varphi}\sim\mathcal{N}(\boldsymbol{0}, \sigma^2\boldsymbol{I})$ models the error in the mocap prediction, where we experimentally set $\sigma=\Delta t$ \hl{based on the assumption that the mocap-estimated acceleration follows a normal distribution with a variance of 1}.
%
%
With the optimized camera position $\boldsymbol{p}_{\mathrm{cam}}$ and its confidence $n$ from the camera tracking module, the state can also be corrected in 30 FPS based on the following correction equation:
\begin{equation}
\label{eq:observation}
    \boldsymbol{p}^{\mathrm{cam}}_{k} = \boldsymbol{p}_{k} + \boldsymbol{p}_{k}^\mathrm{root\rightarrow cam}+\boldsymbol{\psi}_{k},
\end{equation}
where $\boldsymbol{p}_\mathrm{root\rightarrow cam}$ is the position difference between the camera and the root, which can be computed by the estimated human pose using forward kinematics, and $\boldsymbol{\psi}\sim\mathcal{N}(\boldsymbol{0}, \boldsymbol{\varSigma}_{\mathrm{cam}})$ models the noise in the camera tracking, where the covariance matrix $\boldsymbol{\varSigma}_{\mathrm{cam}}$ is computed from the camera position confidence $n$ as:
\begin{equation}
    \boldsymbol{\varSigma}_\mathrm{cam}=\frac{1000}{n+\epsilon}\boldsymbol{I},
\end{equation}
where $\boldsymbol{I}\in\mathbb{R}^{3\times 3}$ is the identity matrix and $\epsilon=10^{-3}$ is used to avoid division by zero.
%
%
Given the prediction and correction equations (Eq.~\ref{eq:kf} and \ref{eq:observation}), we predict the state variables based on the prediction-correction algorithm (details are presented in App.~\ref{app:kalman-filter}).

\subsection{System Calibration}\label{sec:method-initialization}
Here we discuss how to initialize the inertial mocap module and the SLAM module respectively, as well as how to calibrate the coordinate system of the two modules.
%
%
%
Inertial mocap needs a T-pose calibration before capturing the human motion, where the sensor-to-bone rotations and the relative rotation between the IMU inertial frame and the user-defined global frame are determined~\cite{TransPose,DIP,PIP}.
The SLAM module needs a monocular initialization, where the initial map points are triangulated by computing the relative pose between two frames based on the homography or the fundamental matrix~\cite{ORBSLAM}.
\par
In addition to these two initialization steps, we also need to calibrate the extrinsic between the mocap coordinate frame and the SLAM coordinate frame.
We propose an easy-to-perform method, where we require the user to walk in a curve for several seconds, and then the extrinsic can be automatically computed.
Specifically, we express the 
mocap-to-SLAM extrinsic as a similarity transformation in $\mathrm{Sim}(3)$ as the real scale is unknown for monocular SLAM.
During the curve motion, we first independently reconstruct the trajectory of the root by mocap and the camera by SLAM.
Then, given the body poses and fixing the camera on the head, we can align the two trajectories and get the similarity transformation between the mocap coordinate frame and the SLAM coordinate frame.
%
%
%
%

\section{Experiments}
%
%
In this section, we first present the implementation details including the setup of our live system and the datasets used in our experiments (Sec.~\ref{sec:implementation-details}). 
Then, we compare our method with the state-of-the-art techniques in both the research fields of motion capture (mocap) with sparse inertial sensors and SLAM (Sec.~\ref{sec:comparisons}).
Next, we evaluate the key designs and components in our method (Sec.~\ref{sec:evaluations}).
Finally, we discuss our limitations (Sec.~\ref{sec:limitations}).
More results can be found in our supplemental video.

\subsection{Implementation Details}\label{sec:implementation-details}
%
%
%
%
\subsubsection{System Setup}
%
%
%
The system receives 60Hz IMU signals from 6 Xsens Dot~\cite{Xsens} IMUs via Bluetooth and 30Hz RGB images in the resolution of $640 \times 480$ from a smartphone camera via Wi-Fi.
Then, the two signals are synchronized by a user's jumping motion, which is easy to be detected from the two signals.
Our method runs on a laptop with Intel(R) Core(TM) i7-12700H CPU.
Note that we do not need a specific graphic card as our method runs purely on CPU.
%
%
The whole system runs in real time at about 60 FPS.
%
%
%
%
We implement our main method using Pytorch~\cite{pytorch} and Rigid Body Dynamic Library (RBDL)~\cite{RBDL} in python.
The SLAM module is implemented in C++ to achieve real-time performance and is compiled as a shared object library that can be dynamically linked by our main method.
\subsubsection{Datasets}
The datasets used in the experiments include DIP-IMU~\cite{DIP}, AMASS~\cite{AMASS}, TotalCapture~\cite{TotalCapture}, and HPS~\cite{HPS}.
AMASS and DIP-IMU datasets are used to train our mocap networks and we follow~\cite{TransPose, PIP} to use them for training and fine-tuning respectively. 
%
TotalCapture and HPS datasets are used to evaluate our technique.
Since we combine sparse inertial mocap and SLAM, we require synchronized and calibrated visual and inertial data to run our technique.
Thus, for TotalCapture, we synthesize visual data by constructing virtual scenes and putting virtual cameras on a virtual character driven by the mocap data. 
For HPS, we calibrate the inertial and visual signals by the first short clip of each motion sequence and discard those sequences that cannot be calibrated. 
Details about the dataset preprocessing can be found in App.~\ref{app:dataset-preprocessing}.
An overview of the training and evaluation datasets is shown in Tab.~\ref{tab:datasets}.
\begin{table}[t]
\caption{Overview of the datasets. The table shows whether the datasets contain pose, translation, IMU measurements, videos from first-person cameras, and scene meshes/point clouds. "Y/N" indicates the dataset does/does not contain such data. "S" means such data is synthetic.}
\label{tab:datasets}
\begin{minipage}{\columnwidth}
\begin{center}
\resizebox{\linewidth}{!}{
\begin{tabular}{ccccccc}
\toprule
Datasets     & Pose  & Translation & IMU & FP Camera & Scene & Minutes \\ \midrule
AMASS        & Y     & Y     & S   & N   & N   & 1217   \\ 
DIP-IMU      & Y     & N     & Y   & N   & N   & 80     \\
TotalCapture & Y     & Y     & Y   & S   & S   & 49     \\
HPS          & Y*    & Y*    & Y   & Y   & Y   & 180    \\
\bottomrule
\end{tabular}}
\end{center}
\footnotetext{*We use the results of the HPS~\cite{HPS} method as ground truths.}
\end{minipage}
\end{table}

\subsection{Comparisons}\label{sec:comparisons}
\begin{table*}[]
\caption{Comparisons with sparse inertial mocap approaches TransPose~\cite{TransPose}, TIP~\cite{TIP}, and PIP~\cite{PIP} on TotalCapture~\cite{TotalCapture} (categorized by motions) and HPS~\cite{HPS} (categorized by scenes) datasets. The reported numbers are absolute root position errors \hl{averaged over the frames} in meters. For our method, we test it 9 times and report the median error (the top number) and the standard deviation (the bottom number, beginning with "$\pm$"). 
}
\label{tab:tran-cmp}
\resizebox{\linewidth}{!}{
\begin{tabular}{cccccccccccccccc}
\toprule
\multirow{2}{*}{Method} & \multicolumn{5}{c}{TotalCapture}               & & \multicolumn{9}{c}{HPS}                                                   \\ \cmidrule{2-6} \cmidrule{8-16} 
                        & acting & freestyle & rom           & walking & average & & BIB\_AB & BIB\_EG & EG   & Etage6 & GEB  & BIB\_UG & KINO & BIB\_OG & average\\ \midrule
TransPose               & 0.53   & 0.69      & 0.18          & 0.45    & 0.42    & & 6.83    & 5.53    & 2.40 & 2.77   & 1.91 & 4.99    & 2.20 & 4.58    & 4.11   \\
TIP                     & 0.43   & 0.87      & 0.21          & 0.49    & 0.45    & & 2.23    & 3.41    & \textbf{1.43} & 3.87   & 1.38 & 2.92    & 0.89 & 3.89 & 3.00\\
PIP                     & 0.61   & 0.51      & \textbf{0.07} & 0.49    & 0.37    & & 1.26    & 2.59    & 1.89 & 1.78   & 1.35 & 2.49    & 1.50 & 4.81    & 2.75   \\ \hline
\multirow{2}{*}{Ours}   & \textbf{0.28}   & \textbf{0.33}      & 0.10  & \textbf{0.25}    & \textbf{0.22}    & & \textbf{1.23}    & \textbf{1.54}    & {1.83} & \textbf{1.35}   & \textbf{1.17} & \textbf{2.40}    & \textbf{0.87} & \textbf{1.90}    & \textbf{1.70}   \\
                        & $\pm$0.06  & $\pm$0.06 & $\pm$0.02 & $\pm$0.03 & $\pm$0.04  & & $\pm$0.37 & $\pm$0.28 & $\pm$0.37 & $\pm$0.18 & $\pm$0.29 & $\pm$0.49 & $\pm$0.16 & $\pm$0.41 & $\pm$0.34\\
\bottomrule
\end{tabular}}
\end{table*}
\begin{table*}[]
\caption{Comparisons with ORB-SLAM3~\cite{ORBSLAM3} in [M]onocular/[M]onocular-[I]nertial [Off]line/[On]line modes. We calculate camera localization errors in meters and we present the numbers on both the [full] sequences and the [tracked] frames (excluding the tracking-failed frames) for the compared methods.
We test all methods 9 times on each sequence and report the median error (the top number), the standard deviation (beginning with "$\pm$"), and the percentage of the tracked frames (ending with "\%").
If a method fails too many times (>85\%) during evaluation, we mark it as a failure (denoted as "-"). 
}  
\label{tab:loc-cmp}
\begin{minipage}{\textwidth}
\begin{center}
\resizebox{\linewidth}{!}{
\begin{tabular}{ccccccccccccccccc}
\toprule
\multicolumn{2}{c}{\multirow{2}{*}{Method}} & \multicolumn{5}{c}{TotalCapture}              & & \multicolumn{9}{c}{HPS}       \\ \cmidrule{3-7} \cmidrule{9-17} 
      &                         & acting    & freestyle & rom       & walking   & average   & & BIB\_AB   & BIB\_EG   & EG        & Etage6    & GEB       & BIB\_UG   & KINO      & BIB\_OG   & average\\ \midrule    
      &                         & 0.82      & 0.89      & 0.25      & 0.42      & 0.54      & & 8.58      & 12.57     & 8.87      & 5.62      & 1.62      & 9.29      & 4.79      & 7.61      & 8.18      \\ 
      & \multirow{-2}{*}{M-Off} & $\pm$0.44 & $\pm$0.17 & $\pm$0.16 & $\pm$0.46 & $\pm$0.29 & & $\pm$0.92 & $\pm$2.30 & $\pm$2.23 & $\pm$1.62 & $\pm$0.15 & $\pm$1.52 & $\pm$0.52 & $\pm$2.06 & $\pm$1.71 \\ \rowcolor{gray!13} \cellcolor{white}
      &                         & 1.34      & 1.01      & 0.25      & 0.80      & 0.76      & & 8.59      & 13.94     & 8.95      & 9.04      & 1.84      & 9.91      & 5.78      & 7.77      & 9.38      \\ \rowcolor{gray!13} \cellcolor{white}
      & \multirow{-2}{*}{M-On}  & $\pm$0.47 & $\pm$0.17 & $\pm$0.16 & $\pm$0.48 & $\pm$0.30 & & $\pm$0.92 & $\pm$2.02 & $\pm$1.20 & $\pm$1.76 & $\pm$0.24 & $\pm$1.43 & $\pm$0.67 & $\pm$1.19 & $\pm$1.41 \\
      &                         & 10.54     & 4.75      & -         & 1.08      & 4.87      & & -         & 78.32     & -         & -         & -         & 7.23      & -         & -         & 67.64     \\
      & \multirow{-2}{*}{MI-Off}& $\pm$5.48 & $\pm$2.62 & -         & $\pm$1.88 & $\pm$3.24 & & -         & $\pm$136.05 & -       & -         & -         & $\pm$5.98 & -         & -         & $\pm$116.52 \\ \rowcolor{gray!13} \cellcolor{white}
      &                         & 2.08      & 0.95      & -         & 0.92      & 1.33      & & -         & 78.87     & -         & -         & -         & 7.31      & -         & -         & 68.12     \\ \rowcolor{gray!13} \cellcolor{white} \multirow{-8}{*}{\rotatebox[origin=c]{90}{Full}}
      & \multirow{-2}{*}{MI-On} & $\pm$0.56 & $\pm$0.41 & -         & $\pm$1.10 & $\pm$0.84 & & -         & $\pm$136.18 & -       & -         & -         & $\pm$6.23 & -         & -         & $\pm$116.66 \\ 
      \hline
      &                         & 0.35      & \textbf{0.32}      & 0.23      & 0.36      & 0.30      & & 2.96      & 5.65      & 6.67      & 4.43      & \textbf{0.67}      & 4.20      & 0.88      & 8.73      & 5.19      \\
      &                         & $\pm$0.50 & $\pm$0.20 & $\pm$0.28 & $\pm$0.36 & $\pm$0.34 & & $\pm$2.48 & $\pm$3.50 & $\pm$3.11 & $\pm$1.62 & $\pm$0.62 & $\pm$5.57 & $\pm$1.03 & $\pm$3.04 & $\pm$2.75 \\
      & \multirow{-3}{*}{M-Off} & 85.9\%    & 58.9\%    & 85.0\%    & 89.7\%    & 80.7\%    & & 51.8\%    & 62.1\%    & 37.3\%    & 61.1\%    & 55.5\%    & 22.5\%    & 61.2\%    & 40.4\%    & 46.5\%    \\ \rowcolor{gray!13} \cellcolor{white}
      &                         & 0.92      & 0.52      & 0.24      & 0.52      & 0.50      & & 3.88      & 9.44      & 6.41      & 7.59      & 1.09      & 3.46      & 2.04      & 10.33     & 7.15      \\ \rowcolor{gray!13} \cellcolor{white}
      &                         & $\pm$0.44 & $\pm$0.29 & $\pm$0.25 & $\pm$0.54 & $\pm$0.37 & & $\pm$1.33 & $\pm$3.35 & $\pm$3.22 & $\pm$2.48 & $\pm$0.94 & $\pm$5.12 & $\pm$1.34 & $\pm$2.08 & $\pm$2.62 \\ \rowcolor{gray!13} \cellcolor{white}
      & \multirow{-3}{*}{M-On}  & 83.1\%    & 62.3\%    & 83.3\%    & 87.6\%    & 79.7\%    & & 51.2\%    & 59.1\%    & 43.7\%    & 58.1\%    & 55.6\%    & 18.1\%    & 62.3\%    & 39.5\%    & 45.2\%    \\
      &                         & 0.32      & 0.39      & -         & 0.58      & 0.53      & & -         & 77.84     & -         & -         & -         & 7.15      & -         & -         & 66.98     \\
      &                         & $\pm$0.07 & $\pm$0.11 & -         & $\pm$0.60 & $\pm$0.48 & & -         & $\pm$132.20 & -       & -         & -         & $\pm$5.73 & -         & -         & $\pm$112.77 \\
      & \multirow{-3}{*}{MI-Off}& 14.5\%    & 5\%       & 0\%       & 59.7\%    & 17.1\%    & & 0\%       & 88.6\%    & 0\%       & 0\%       & 0\%       & 15.9\%    & 0\%       & 0\%       & 18.7\%    \\ \rowcolor{gray!13} \cellcolor{white}
      &                         & 0.41      & 0.37      & -         & 0.59      & 0.54      & & -         & 78.35     & -         & -         & -         & 7.23      & -         & -         & 67.44     \\ \rowcolor{gray!13} \cellcolor{white}
      &                         & $\pm$0.08 & $\pm$0.11 & -         & $\pm$0.62 & $\pm$0.49 & & -         & $\pm$132.23 & -       & -         & -         & $\pm$5.97 & -         & -         & $\pm$112.84 \\ \rowcolor{gray!13} \cellcolor{white} \multirow{-12}{*}{\rotatebox[origin=c]{90}{Tracked}} 
      & \multirow{-3}{*}{MI-On} & 12.1\%    & 7.6\%     & 0\%       & 57.7\%    & 16.7\%    & & 0\%       & 88.6\%    & 0\%       & 0\%       & 0\%       & 15.9\%    & 0\%       & 0\%       & 18.7\%    \\ \hline
      &                         & \textbf{0.29}      & {0.35}      & \textbf{0.13}      & \textbf{0.25}      & \textbf{0.24}      & & \textbf{1.25}      & \textbf{1.53}      & \textbf{1.81}       & \textbf{1.34}      & {1.18}      & \textbf{2.39}      & \textbf{0.86}      & \textbf{1.90}      & \textbf{1.69}\\
      & \multirow{-2}{*}{Ours}  & $\pm$0.06 & $\pm$0.06 & $\pm$0.02 & $\pm$0.04 & $\pm$0.04 & & $\pm$0.37 & $\pm$0.28 & $\pm$0.36 & $\pm$0.18 & $\pm$0.29 & $\pm$0.49 & $\pm$0.16 & $\pm$0.41 & $\pm$0.33\\
\bottomrule   
\end{tabular}}
\end{center}
\end{minipage}
\end{table*}
\begin{figure*}[t]
  \includegraphics[width=\linewidth]{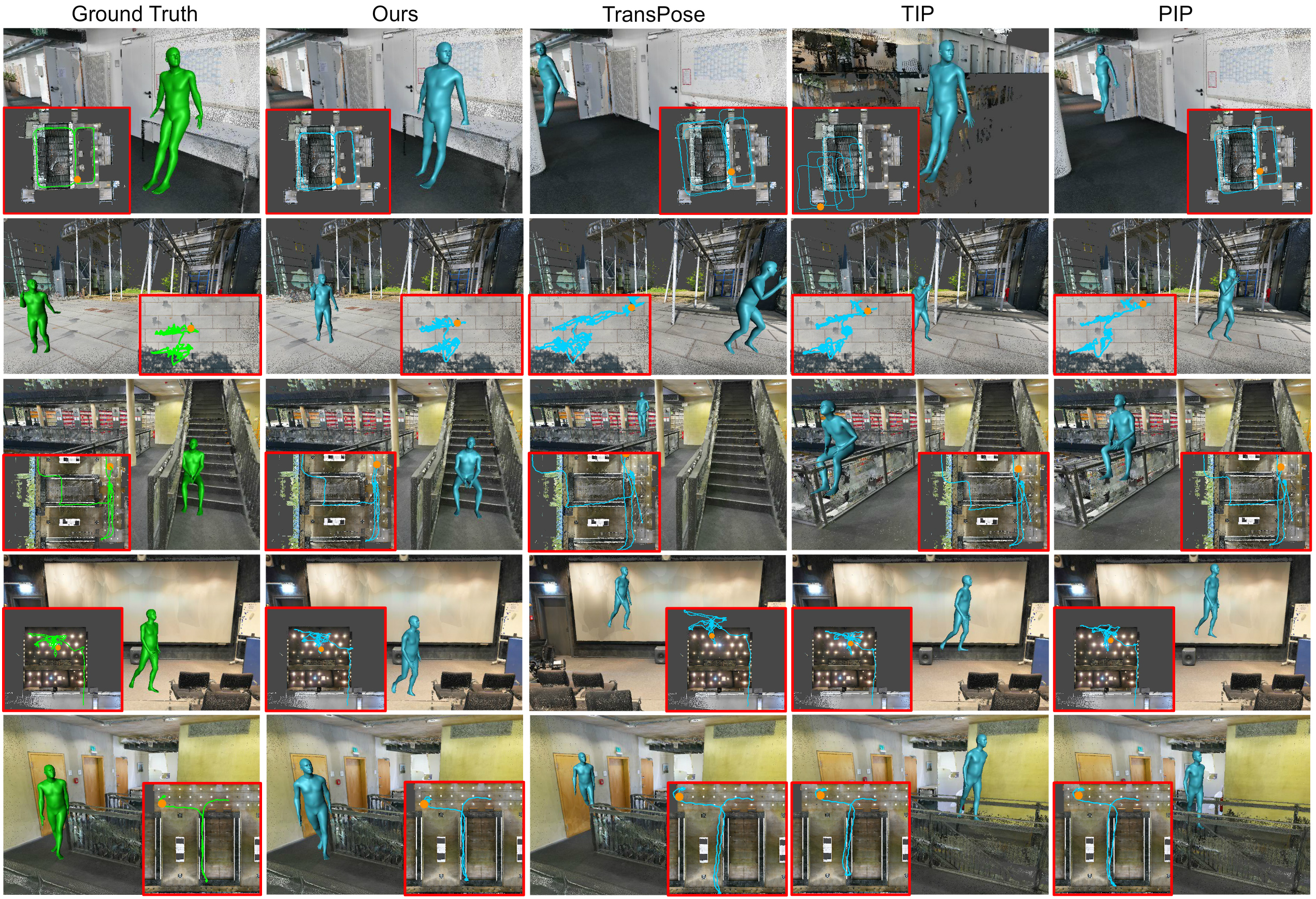}
  \caption{Qualitative comparisons on human motion capture with TransPose~\cite{TransPose}, TIP~\cite{TIP}, and PIP~\cite{PIP} on HPS~\cite{HPS} dataset. We show the ground-truth humans in green and the estimated humans in blue. The movements of the human are also plotted in the top view. The orange point in the top view denotes the current global position of the human.
  }
  \label{fig:mocap-cmp}
\end{figure*}
Our work simultaneously captures human motion and performs localization and mapping from sparse IMUs and a wearable camera.
Since there is no existing technique that solves these two tasks simultaneously, we compare our method with the state-of-the-art methods of both fields, \textit{i.e.}, TransPose~\cite{TransPose}, TIP~\cite{TIP}, and PIP~\cite{PIP} for mocap from sparse IMUs, ORB-SLAM3~\cite{ORBSLAM3} for monocular/monocular-inertial SLAM.
\subsubsection{Comparisons with Sparse Inertial Mocap}
%
%
%
%
We first quantitatively compare the methods using \textit{absolute root position error}, \textit{i.e.}, the average root position error of all frames, on TotalCapture~\cite{TotalCapture} and HPS~\cite{HPS} datasets, with the root's position and orientation aligned with the ground truth at the first frame.
The results are shown in Tab.~\ref{tab:tran-cmp}, where we achieve as large as 41\% and 38\% improvement on TotalCapture and HPS datasets respectively over the state-of-the-art methods.
Our method significantly outperforms previous works in different scenes including small/large indoor/outdoor environments and for different human motions.
%
%
It demonstrates that our method successfully and robustly leverages the camera to reduce the translation drift in inertial mocap.
One special case is the "rom" motions in the TotalCapture dataset, which contain very few global movements of humans, making it difficult to correctly initialize the SLAM system and determine the scale factor in our method, and such errors reflect in the results.
Another special case is the "EG" scene in the HPS dataset, where TIP outperforms our method on the root position error.
However, TIP still suffers from large translation drifts most of the time, which is reflected in other sequences.
While our method is more robust and reduces 43\% errors on average on the HPS dataset compared with TIP.
%
%
Note that since all the methods assume a mean body shape, we rescale the estimated translation with the ground-truth leg lengths when evaluating the position error.
Also note that the three pure mocap algorithms (TransPose, TIP, and PIP) are \textit{deterministic}, \textit{i.e.}, there is no randomness in the results.
While our method is \textit{stochastic} (due to the RANSAC algorithm and multi-threading in the SLAM part).
Thus, following previous SLAM works, we test our method 9 times for each sequence and report the median error as well as the standard deviation.
\par
We also present qualitative mocap comparison results in Fig.~\ref{fig:mocap-cmp}.
Due to the error accumulation in inertial mocap, previous approaches suffer from translation drifts.
Thus, as shown in the figure, they estimate wrong human positions in the scene as time goes by.
By incorporating SLAM in the human mocap, our method largely reduces the translation drifts during human motion, and the global position of the human is the most accurate. 
Note that the beginning human global position and orientation are aligned with the ground truth in all the methods, and the results are selected after a period of human motion (shown as the trajectory lines in the figure).
All methods are not aware of the ground-truth scene.
\par
%
%
Regarding the comparisons on human pose estimation, we would like to note that involving SLAM in mocap majorly contributes to determining the global positions, rather than the human poses.
Thus, we present the pose estimation results in App.~\ref{app:pose-comparisons}, where we achieve similar pose estimation accuracy compared with the state-of-the-art mocap methods.
\begin{table*}[]
\caption{Comparisons on mapping accuracy with [M]onocular/[M]onocular-[I]nertial ORB-SLAM3~\cite{ORBSLAM3}. For each of the three scenes, we synthesize monocular videos based on the human motions in TotalCapture~\cite{TotalCapture} dataset. We test each method 9 times and report the median map point errors (the top number) and the standard deviations (the bottom number, beginning with "$\pm$") in meters. If a method fails too many times (>85\%) during evaluation, \textit{i.e.}, no map point is reconstructed, we mark it as a failure (denoted as "-").}
\label{tab:map-cmp}
\resizebox{\linewidth}{!}{
\begin{tabular}{cccccccccccccccccc}
\toprule
\multirow{2}{*}{Method} & \multicolumn{5}{c}{Japan Office}  & & \multicolumn{5}{c}{Flooded Grounds} & & \multicolumn{5}{c}{SciFi Warehouse} \\ \cmidrule{2-6} \cmidrule{8-12} \cmidrule{14-18}
                           & acting        & freestyle     & rom           & walking       & average       & 
                           & acting        & freestyle     & rom           & walking       & average       & 
                           & acting        & freestyle     & rom           & walking       & average       \\ \midrule
                           & 0.88          & 1.10          & 1.20          & 0.60          & 0.95          &
                           & 2.89          & 2.06          & 3.10          & 1.36          & 2.31          & 
                           & 0.40          & 0.43          & \textbf{0.44} & 0.35          & 0.41          \\
\multirow{-2}{*}{SLAM (M)} & $\pm$0.26     & $\pm$0.76     & $\pm$1.69     & $\pm$0.30     & $\pm$0.76     & 
                           & $\pm$1.52     & $\pm$1.30     & $\pm$1.65     & $\pm$1.31     & $\pm$1.44     & 
                           & $\pm$0.03     & $\pm$0.13     & $\pm$0.05     & $\pm$0.02     & $\pm$0.06     \\ \rowcolor{gray!13}
                           & 0.66          & 0.54          & -             & 0.61          & 0.61          &
                           & -             & -             & -             & 0.90          & -          & 
                           & 0.49          & 0.40          & -             & 0.51          & 0.48          \\ \rowcolor{gray!13}
\multirow{-2}{*}{SLAM (MI)}& $\pm$0.83     & $\pm$1.50     & -             & $\pm$0.49     & $\pm$0.79     & 
                           & -             & -             & -             & $\pm$0.77     & -     & 
                           & $\pm$0.10     & $\pm$0.13     & -             & $\pm$0.41     & $\pm$0.28     \\
                           & \textbf{0.32} & \textbf{0.49} & \textbf{0.72} & \textbf{0.24} & \textbf{0.45} &
                           & \textbf{0.94} & \textbf{1.49} & \textbf{1.75} & \textbf{0.80} & \textbf{1.23}  & 
                           & \textbf{0.24} & \textbf{0.37} & \textbf{0.44} & \textbf{0.18} & \textbf{0.31} \\
\multirow{-2}{*}{Ours}     & $\pm$0.09     & $\pm$0.09     & $\pm$0.23     & $\pm$0.06     & $\pm$0.12     & 
                           & $\pm$0.31     & $\pm$0.70     & $\pm$1.13     & $\pm$0.19     & $\pm$0.56     & 
                           & $\pm$0.03     & $\pm$0.10     & $\pm$0.17     & $\pm$0.02     & $\pm$0.08    \\
\bottomrule
\end{tabular}}
\end{table*}
\begin{figure*}[t]
  \includegraphics[width=\linewidth]{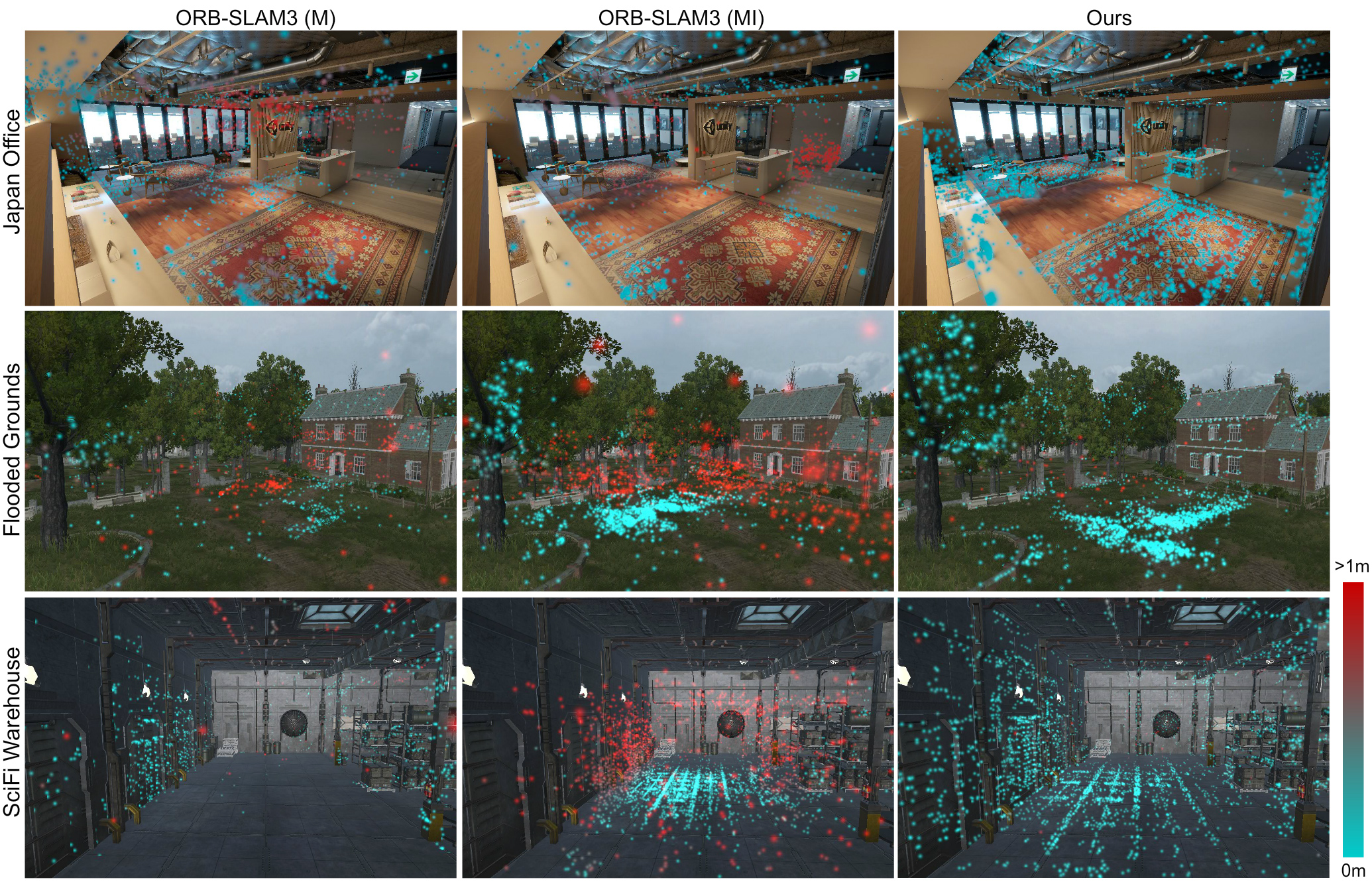}
  \caption{Qualitative mapping comparisons with [M]onocular/[M]onocular-[I]nertial ORB-SLAM3~\cite{ORBSLAM3}. We show the reconstructed map points in a walking sequence for different methods. Color encodes the map point error. For monocular SLAM, we calculate the map scale from the ground truth.
  }
  \label{fig:map}
\end{figure*}
\subsubsection{Comparisons with SLAM}
We compare our method with ORB-SLAM3~\cite{ORBSLAM3}, which is a state-of-the-art technique in visual and visual-inertial SLAM.
We run two modes in ORB-SLAM3: \textit{monocular (M)} mode which takes RGB stream as input, and \textit{monocular-inertial (MI)} mode which uses synchronized RGB images and IMU measurements.
In each mode, we output both the \textit{online (On)} localization results from the SLAM front-end at running frame rate, and the \textit{offline (Off)} localization results from the SLAM back-end after running the whole sequence.
%
%
The offline results are usually better than the online results as they have been optimized in the bundle adjustment in the back end, but they cannot be obtained online in real time.
%
%
For our system, we always calculate real-time camera positions online from the estimated human pose and translation.
%
%

%
\paragraph{Localization Comparisons}
We quantitatively compare different solutions on TotalCapture~\cite{TotalCapture} and HPS~\cite{HPS} datasets by \textit{absolute camera position error}.
%
%
%
%
%
%
For a fair comparison, we should calculate errors on the \textit{full} sequences for all methods.
However, due to fast human motions, the SLAM system may often fail to track the camera, which will lead to very large errors.
To this end, we also report the errors calculated on the successfully \textit{tracked} frames for the SLAM solutions, while for our method we consistently report the errors on full frames as we do not suffer the failure.
%
%
%
%
%
%
%
%
\par
The comparison results are shown in Tab.~\ref{tab:loc-cmp}.
Among different settings, online monocular-inertial SLAM (MI-on) is the closest to our method, as both methods leverage inertial sensors and run online.
Our method consistently outperforms MI-on on both full sequences and tracked frames, achieving 82\% and 56\% improvement respectively.
Offline monocular SLAM (M-Off) occasionally achieves lower errors on tracked frames than our method.
However, it runs offline, and up to 36\% human motions fail to be tracked. 
Furthermore, our real-time method still outperforms it on average on both datasets. 
Overall, our method achieves lower localization errors and lower standard deviations on both datasets, which reflects that the combination of mocap and SLAM greatly improves both the accuracy and robustness of SLAM systems.
Note that for monocular SLAM (M-Off, M-On), we calculate the scale of the estimated trajectory by the ground truth as monocular SLAM has an inherent ambiguity here.
Our method and MI do not suffer this ambiguity.
\par
We would like to list two further advantages of our system over monocular/monocular-inertial ORB-SLAM3.
%
%
\textit{1)} \textit{Less sensitive to calibration errors.} Readers may find that monocular-inertial ORB-SLAM3 performs well on TotalCapture but nearly fails on HPS. This is mainly because HPS contains more calibration errors on camera-IMU extrinsic and the frequency of IMU signals is lower (see dataset preprocessing in App.~\ref{app:dataset-preprocessing}). Nevertheless, our method performs well on both datasets with the same extrinsic calibration. 
\textit{2)} \textit{No failures.} Pure SLAM systems often fail to track the camera, which is reflected in the low percentage of tracked frames. This could happen when the human moves quickly or just turn around. By combining mocap and SLAM, our system always gives an estimation based on the mocap result even when the visual tracking is lost. 

\paragraph{Mapping Comparisons}
In addition to localization comparisons, we also compare the mapping accuracy with ORB-SLAM3~\cite{ORBSLAM3} by extracting the reconstructed 3D map points at the end of each sequence.
We evaluate \textit{map point error}, \textit{i.e.}, the average distance between each reconstructed map point and the nearest scene point, on three different scenes\footnote{The three scenes include Japan Office (\url{http://aec.unity3d.jp/}), Flooded Grounds (\url{https://assetstore.unity.com/packages/3d/environments/flooded-grounds-48529}), and SciFi Warehouse (\url{https://assetstore.unity.com/packages/3d/environments/sci-fi/sci-fi-construction-kit-modular-159280}). All the scenes are free assets and can be accessed from Unity Asset Store.}.
The quantitative evaluation results are shown in Tab.~\ref{tab:map-cmp}.
Our method consistently outperforms monocular/monocular-inertial ORB-SLAM3 in different scenes and for different human motions, demonstrating that the deep fusion of mocap and SLAM not only helps with human localization but also enhances the mapping accuracy largely.
We also qualitatively show the mapping results of the three scenes in Fig.~\ref{fig:map}.
Our method accurately reconstructs sparse environment points with much fewer noises compared with the SLAM methods.
It also robustly handles unconstrained outdoor large environments ("Flooded Grounds"). 
This is attributed to the close fusion of inertial mocap and SLAM, which involves mocap priors in the mapping.
Readers may notice that the reconstructed points are sparser in "Flooded Grounds" and has a larger error than the indoor scenes.
This is because the outdoor environment has fewer visual feature points, and reconstructing a far point of a large scene is more difficult and prone to errors in multi-view 3D reconstruction.
Besides, the mapping errors on "rom" motions are larger compared with other human motions.
This is because the "rom" motions contain very few global movements, making it difficult to reconstruct the map due to the high similarity of camera views.

\subsection{Evaluations}\label{sec:evaluations}
\begin{table}[]
\caption{Run-time performance on an Intel(R) Core(TM)
i7-12700H CPU on a laptop. We show the time costs in milliseconds when processing one frame.}
\label{tab:runtime-performance}
\resizebox{\linewidth}{!}{
\begin{tabular}{cccccc}
\toprule
sync signals & motion capture & cam tracking & tran update & mapping \\ \midrule
4.7 & 5.3 & 12.8 & 0.2 & 97.1 \\
\bottomrule
\end{tabular}}
\end{table}
\subsubsection{Performance}
We first evaluate the run-time performance of our method and the results are shown in 
Tab.~\ref{tab:runtime-performance}.
Note that our system receives IMU signals in 60 FPS and color images in 30 FPS.
To process a frame with inertial signals, we need to synchronize multiple IMU signals, perform inertial mocap, and update the translation, which takes about 10.2 milliseconds.
To process a frame with both inertial and visual signals, we need additional camera tracking, which takes 23 milliseconds in total.
Therefore, in 1 second, we can process 30 pure-inertial frames and 30 visual-inertial frames in total, \textit{i.e.}, our system can leverage all the available input signals and output pose and translation in 60 FPS.
While the mapping (and loop closing) takes much more time than the tracking, we run it only on \textit{keyframes} and as a background task through multi-threading to avoid additional time costs.
Thus, it does not affect our real-time 60 FPS performance.

\subsubsection{Effectiveness}
\begin{table}[]
\caption{Ablation study on the methods of combining mocap and SLAM. We report the median root position errors and the standard deviations in meters in 9 tests on the TotalCapture~\cite{TotalCapture} dataset.}
\label{tab:ablation}
\resizebox{\linewidth}{!}{
\begin{tabular}{cccccc}
\toprule
\multirow{2}{*}{Method}                & \multicolumn{5}{c}{TotalCapture}                                     \\ \cmidrule{2-6}
                                       & acting     & freestyle    & rom         & walking     & average      \\ \midrule
                                       & 0.72       & 0.64         & 0.12        & 0.57        & 0.46         \\
\multirow{-2}{*}{inertial tracking}    & $\pm$0     & $\pm$0       & $\pm$0      & $\pm$0      & $\pm$0        \\ \rowcolor{gray!13}
                                       & 1.34       & 1.01         & 0.25        & 0.80        & 0.76         \\ \rowcolor{gray!13}
\multirow{-2}{*}{monocular SLAM}       & $\pm$0.47  & $\pm$0.17    & $\pm$0.16   & $\pm$0.48   & $\pm$0.30    \\ \hline
                                       & 1.90       & 0.74         & 0.13        & 0.98        & 0.82         \\
\multirow{-2}{*}{fuse as init}         & $\pm$0.36  & $\pm$0.17    & $\pm$0.03   & $\pm$0.23   & $\pm$0.17    \\ \rowcolor{gray!13}
                                       & 1.13       & 0.66         & 0.12        & 0.92        & 0.63         \\ \rowcolor{gray!13}
\multirow{-2}{*}{fused tracking}       & $\pm$0.35  & $\pm$0.07    & $\pm$0.01   & $\pm$0.26   & $\pm$0.15    \\
fused tracking\&mapping               & 0.29       & 0.34         & \textbf{0.09}        & 0.28        & 0.23         \\
\small{(w/o confidence)}                       & $\pm$0.13  & $\pm$0.08    &  $\pm$0.03  & $\pm$0.08   & $\pm$0.07    \\ \rowcolor{gray!13}
                                       & \textbf{0.28}       & \textbf{0.33}         & 0.10        & \textbf{0.25}        & \textbf{0.22}        \\ \rowcolor{gray!13}
\multirow{-2}{*}{Ours}                 & $\pm$0.06  & $\pm$0.06    & $\pm$0.02   & $\pm$0.03   & $\pm$0.04    \\

\bottomrule
\end{tabular}}
\end{table}
We demonstrate the effectiveness of our method by conducting ablative experiments on different combining methods of inertial mocap and SLAM.
Specifically, we evaluate 
\textit{1) inertial tracking}, the baseline mocap method where we estimate the translation only from 6 IMUs using the inertial motion capture module, which is similar to PIP~\cite{PIP} but the flat ground assumption and the force calculation are removed; 
\textit{2) monocular SLAM}, the baseline SLAM method where the translation is estimated by tracking the monocular video purely;
%
%
\textit{3) fuse as init}, \hl{where we initialize the camera pose in monocular SLAM by the mocap-derived values, while removing all the mocap terms in the optimization, and use the body translation updater to get the root translation from the camera. In other words, it is the same as the standard monocular SLAM except that the mocap-derived camera pose is used to initialize the camera tracking optimization, replacing the original initialization scheme based on the constant-speed assumption;}
\textit{4) fused tracking}, where we fuse mocap and SLAM in the camera tracking, \textit{i.e.}, using the proposed mocap-constrained camera tracking;
\textit{5) fused tracking\&mapping}, where we fuse mocap and SLAM in both camera tracking and mapping/loop closing, which is the same as our full method except that the mocap-related map point confidences are not used in the mocap-aware bundle adjustment during optimization.
We evaluate the root position errors on the TotalCapture~\cite{TotalCapture} dataset, and the results are shown in Tab.~\ref{tab:ablation}
Due to the translation drifts in pure \textit{1) inertial tracking} and the frequent tracking failure in pure \textit{2) monocular SLAM}, the two baselines have relatively large errors (Row 1\&2).
%
%
We then show that a naive fusion of mocap and SLAM by \textit{3) fuse as init} will not give good results (Row 3).
With the \textit{4) fused tracking}, the system performs slightly better (Row 4), because the mocap constraints help to reduce the errors caused by keypoint mismatches in the camera tracking.
Nevertheless, the shallow fusion of mocap and SLAM in the camera tracking still can not outperform the mocap baseline.
This is because, though the camera tracking accuracy \textit{w.r.t the map} is improved, the map itself is often poorly reconstructed (\textit{e.g.}, in a wrong scale) due to the unawareness of mocap.
Then, tracking the camera in an inaccurate map will lead to poor localization results, which further degenerate the mapping in a positive-feedback loop.
%
%
To this end, we propose \textit{5) fused tracking\&mapping} (Row 5), which shows a large enhancement in tracking accuracy and robustness compared with the baselines.
We attribute this to the mocap-aware bundle adjustment algorithm, where the map and keyframe camera poses are jointly optimized with mocap prior, which greatly enhances the mapping accuracy.
Furthermore, with the proposed mocap-related map point confidence (Row 6), \hl{the standard deviation of multiple experiments is reduced by 43\%, reflecting that the uncertainty in the tracking is largely reduced. On the other hand, the overall accuracy is further improved slightly.}
%
This shows the effectiveness of the deep fusion of mocap and SLAM in both front-end tracking and back-end mapping, as well as the proposed mocap-related map point confidence.
%
%
%
%
%
%


\subsection{Limitations}\label{sec:limitations}
\paragraph{Online Loop Closure}
In our system, visual information could help to reduce the drift in inertial mocap as we did in Sec.~\ref{sec:method-translation-updater}, \hl{but there are still remaining drifts and loop closing is applied to handle them.}
However, loop closure requires modifying the camera pose of previous frames when a loop is detected, which is not consistent with an online mocap system whose results of previous frames have already been exported and cannot be modified. 
So, in our online system, even though we can leverage the loop closure component in SLAM, we have to teleport the user to its correct position to reduce the drift rather than generate a smooth trajectory.
\hl{
\paragraph{Body Shape and Scene Scale}
By aligning the mocap trajectory and the SLAM trajectory during initialization, the scene scale is consistent with the body shape scale. However, since we assume mean shape rather than measure the true shape scale of the user, our reconstructed scene scale may be slightly different from the real one. This problem could be handled by measuring the bone lengths and scaling the mocap output accordingly.
\paragraph{Calibration Errors}
To obtain optimal results, data synchronization and system calibration should be carefully performed. In our implementation, we need a jumping motion for sensor synchronization, a still T pose for sensor-to-bone calibration, and a curve movement for global frame alignment. To ensure the calibration is well done, we check the curve trajectory alignment error to determine whether the calibration is successful. Typically, a successful calibration takes about 20 seconds. These steps are required because a bad calibration will lead to large inconsistencies between mocap and SLAM estimations, yielding poor results.
\paragraph{Degenerated Cases}
Our system is robust to a few dynamic objects by leveraging the human prior (\textit{e.g.}, playing ping-pong in the supplemented video, where the second person is always moving in front of the head-mounted camera), but the system will degenerate into inertial mocap or become less accurate if the scene is too dynamic or sparsely textured. Besides, in extreme cases where the human shakes the head fast or performs very weird poses, our system may generate wrong scene points as the mocap fails.
}
\section{Conclusion} 
This is the first work that combines inertial mocap with SLAM to achieve simultaneously human motion capture (mocap), localization, and mapping in real time.
The system is still lightweight as it only requires sparse body-mounted sensors, including 6 inertial measurement units (IMUs) and a monocular phone camera.
For online tracking, the mocap and SLAM densely exchange information by coupled optimizations and Kalman filter techniques, leading to more robust and accurate localization.
For the back-end optimization, both bundle adjustment and loop closure in SLAM benefit from the inertial mocap and the localization errors can be further reduced.

\begin{acks}
This work was supported by the National Key R\&D Program of China (2018YFA0704000), the NSFC (No.62021002), Beijing Natural Science Foundation (M22024), and the Key Research and Development Project of Tibet Autonomous Region (XZ202101ZY0019G). 
This work was also supported by THUIBCS, Tsinghua University, and BLBCI, Beijing Municipal Education Commission. 
This work was partially supported by the ERC consolidator grant 4DReply (770784). 
The authors would like to thank Wenbin Lin, Zunjie Zhu, Guofeng Zhang, and Haoyu Hu for their extensive help on the experiments and live demos.
The authors would also like to thank Ting Shu, Shuyan Han, and Kelan Liu for their help on this project.
Feng Xu is the corresponding author.
\end{acks}

\bibliographystyle{ACM-Reference-Format}
\bibliography{cv}

\appendix
\section{Dataset Preprocessing}\label{app:dataset-preprocessing}
\begin{figure}[t]
  \includegraphics[width=\linewidth]{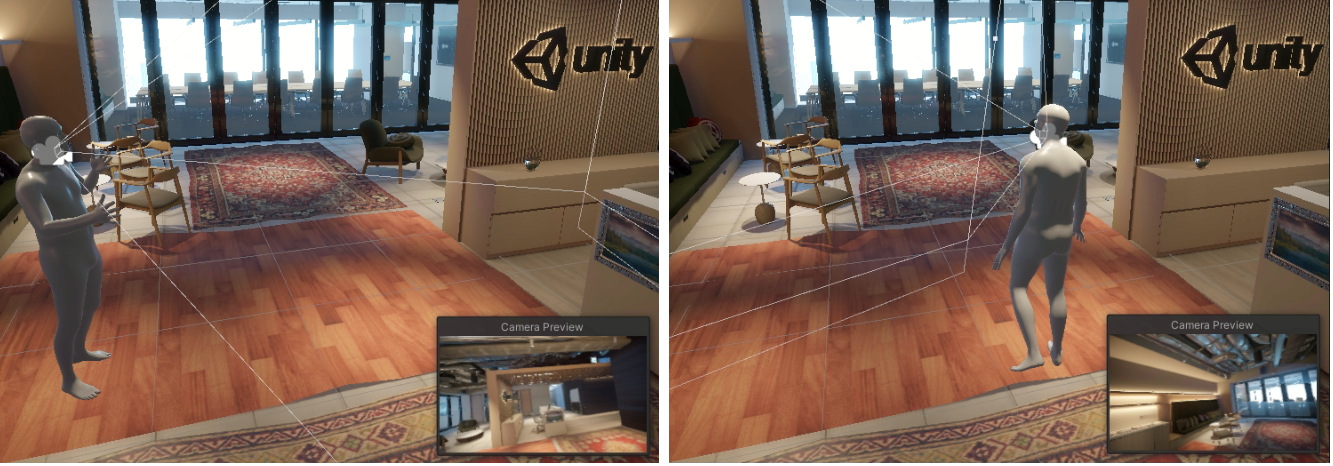}
  \caption{We synthesize first-person videos for the TotalCapture dataset. The pictures at the bottom right corner are the synthesized first-person views.}
  \label{fig:syn_cam}
\end{figure}

In this section, we explain the details of our preprocessing for the datasets, including AMASS~\cite{AMASS}, DIP-IMU~\cite{DIP}, TotalCapture~\cite{TotalCapture}, and HPS~\cite{HPS}.
Since these datasets are recorded for different purposes and have different usages in our experiments, we process each dataset differently.
%
%
\textit{1) AMASS.} This dataset is used only in training. Thus, we use ground-truth pose and translation, and synthesize 60Hz acceleration and orientation measurements from the human motion, following \cite{TransPose, PIP}.
\textit{2) DIP-IMU.} This dataset contains real inertia measurements but no human translation. Thus, we use it in the network fine-tuning step following \cite{TransPose, PIP}.
\textit{3) TotalCapture.} This dataset is used in the evaluation and contains character pose, translation, and IMUs, but no first-person videos. 
Thus, we transfer the ground-truth pose and translation to a SMPL~\cite{SMPL} human model and put the model in virtual scenes. 
With a virtual camera on the model's head, first-person videos are synthesized for this dataset (see Fig.~\ref{fig:syn_cam}). 
Such videos are used to run our method during evaluation, and the virtual scenes are used to validate the mapping accuracy. 
The camera-head relative pose is fixed when synthesizing the first-person videos, and is assumed known in the experiments.
Besides, visual-inertial SLAM assumes known IMU noise parameters (IMU intrinsic), which we assign experimentally in the values that produce the best results, and known camera-IMU relative pose (camera-IMU extrinsic), which we calculate from T-pose calibration. 
Note that the intrinsic and extrinsic are only used by visual-inertial SLAM, while our method does not need them.
%
%
\textit{4) HPS.} This dataset is recorded in large scenes and provides both real inertial measurements and first-person videos. 
%
HPS also estimates character pose and translation by combining dense inertial motion capture and scene-based localization, which are used as the ground truths for our evaluations. 
To evaluate different methods on the dataset, we need three extrinsics: imu-to-bone rotation, camera-to-head rotation, and camera-to-imu rotation, which are not provided in this dataset.
We first use the beginning 150 frames of each sequence to perform IMU-to-bone calibration, which determines imu-to-bone rotations.
Then, the camera-to-head rotation is calculated as the mean rotation between the head orientations measured by the IMU and the camera orientations measured by scene-based localization in the first 60 frames. 
Note that sequences with too much false camera localization in the beginning 60 frames or have evidently wrong IMU-camera synchronization are discarded. 
Finally, the camera-to-imu rotation is computed by multiplying the camera-to-head rotation and the head-to-imu rotation.
To perform visual-inertial SLAM on this dataset, we synthesize the IMU angular velocities and local accelerations from the sensor orientation and free acceleration signals as they are not provided in this dataset.
Mathematically, the sensor-local angular velocity $\boldsymbol{\omega}_S$ and acceleration $\boldsymbol{a}_S$ are calculated as:
\begin{equation}
    \boldsymbol{\omega}_S(t) = \frac{1}{\Delta t}\mathrm{Log}(\boldsymbol{R}_{IS}(t)^{-1}\boldsymbol{R}_{IS}(t+1)),
\end{equation}
\begin{equation}
    \boldsymbol{a}_S(t) = \boldsymbol{R}_{IS}(t)^{-1}(\boldsymbol{a}_I(t) - \boldsymbol{g}_I),
\end{equation}
where $\boldsymbol{R}_{IS}(t)$ and $\boldsymbol{a}_I(t)$ are the sensor orientation and free acceleration in the global inertial frame at frame $t$ respectively, $\Delta t$ is the time interval between two frames, $\boldsymbol{g}_I$ is the gravity in the global inertial frame, and the logarithm map $\mathrm{Log}:\mathrm{SO(3)}\rightarrow\mathbb{R}^3$ maps from the Lie group to the vector space.
This results in 30Hz inertia measurements, the highest frequency of signals we can acquire from the provided data.
The visual-inertial SLAM also needs the IMU intrinsic, which is assigned experimentally, and the camera-IMU extrinsic, which is assigned with the camera-to-imu rotation and zero relative position.
%

%
\begin{table}[]
\caption{Comparisons on pose accuracy with TransPose~\cite{TransPose}, TIP~\cite{TIP}, and PIP~\cite{PIP} on TotalCapture~\cite{TotalCapture} dataset. We report the mean per joint position error (MPJPE) in millimeters with the root position aligned.}
\label{tab:pose-cmp}
\resizebox{0.9\linewidth}{!}{
\begin{tabular}{cccccc}
\toprule
\multirow{2}{*}{Method} & \multicolumn{5}{c}{TotalCapture}    \\ \cmidrule{2-6}
             & acting & freestyle & rom   & walking & average \\ \midrule
TransPose    & 62.9   & 89.3     & 55.6  & 55.0    & 63.6    \\
TIP          & \textbf{56.9}   & \textbf{79.4}     & 56.0  & 49.2    & 59.3\\
PIP          & 60.1   & 82.1     & 45.7  & \textbf{46.6}    & 56.2    \\
Ours         & 59.6   & 82.5      & \textbf{44.9}  & 47.6    & \textbf{56.1}    \\
\bottomrule
\end{tabular}}
\end{table}
\section{Pose Comparisons with Inertial Mocap}\label{app:pose-comparisons}
We conduct pose comparisons with sparse inertial mocap methods TransPose~\cite{TransPose}, TIP~\cite{TIP}, and PIP~\cite{PIP} using mean per joint position error (MPJPE) with the root positions aligned with the ground truth.
The results are shown in Tab.~\ref{tab:pose-cmp}.
We achieve comparable pose estimation accuracy with the state-of-the-art mocap methods.
We would like to note that introducing SLAM in the inertial mocap majorly helps with improving the translation estimation accuracy rather than the pose in our algorithm.
\section{Modification over PIP in mocap}\label{app:method-motion-capture}
The motion capture module is designed following~\cite{PIP}.
We use the same neural kinematics estimator but a different motion optimizer as discussed in Sec.~\ref{sec:method-motion-capture}.
Following the notations in~\cite{PIP}, our new motion optimizer is defined as:
\begin{equation}\label{eq:qp}
\begin{array}{rl}
    \mathop{\arg\min}\limits_{\ddot{\boldsymbol q}}&\|\ddot{\boldsymbol q}_{3:} - \ddot{\boldsymbol \theta}_\mathrm{des}\|^2+ \|\boldsymbol J\ddot{\boldsymbol q} + \dot{\boldsymbol J}\dot{\boldsymbol q} - \ddot{\boldsymbol r}_\mathrm{des}\|^2\\
    \mathrm{s.t.}&\dot{\boldsymbol r}_\mathrm{j}(\ddot{\boldsymbol q})\in \mathcal{C}.\\
\end{array}
\end{equation}
The only optimizable variable is the generalized acceleration $\ddot{\boldsymbol{q}}$ of the human pose and translation, while we do not consider the forces.
The optimization solves the acceleration that best reproduces the kinematically estimated human pose and joint velocity leveraging the dual PD controller.
The contact constraints $\mathcal{C}$ is defined by:
\begin{equation}
  \begin{aligned}
  &\mathcal{C}_j = \{\dot{\boldsymbol r}_j\in \mathbb{R}^3||\dot{\boldsymbol r}_j^x| \le \sigma, |\dot{\boldsymbol r}_j^z| \le \sigma\},\\
  &\mathcal{C} = \{[\dot{\boldsymbol r}_1\cdots\dot{\boldsymbol r}_{n_\mathrm{j}}]\in \mathbb{R}^{3n_\mathrm{j}}|\dot{\boldsymbol r}_j\in\mathcal{C}_j,j=1,2,\cdots,n_\mathrm{j}\},
  \end{aligned}
\end{equation}
where we remove the vertical velocity constraints of the original method as we do not assume a known ground (we allow the capture in the free 3D space).
Besides, we only consider the kinematically estimated contact probabilities during the contact determination, where a foot with a probability larger than $0.5$ is classified as in contact and vice versa. 
Our new optimization can be efficiently solved using quadratic programming algorithms.
Readers are referred to~\cite{PIP} for the notations and more details.
\section{Nonlinear Optimizations in SLAM}\label{app:nonlinear-optimizations}
To accelerate our algorithm, we compute the partial Jacobians analytically for the key optimizations, including the mocap-constrained camera tracking (Eq.~\ref{eq:tracking}) and the mocap-aware bundle adjustment (Eq.~\ref{eq:ba}). We present the details in the following.
\paragraph{Parameterization}
As we express the camera orientation $\boldsymbol{R}\in\mathrm{SO}(3)$ in the over-parameterized matrix representation that lies in a non-Euclidean space, we optimize it by adding a perturbation $\Delta \boldsymbol{\theta}\in\mathbb{R}^3$ around the current value, which uses a minimal representation for 3D rotations.
We define the new "adding" operator $\oplus$ between a rotation matrix and a perturbation as:
\begin{equation}
    \boldsymbol{R}\oplus\Delta\boldsymbol{\theta}=\boldsymbol{R}\mathrm{Exp}(\Delta\boldsymbol{\theta}),
\end{equation}
where the exponential map $\mathrm{Exp}:\mathbb{R}^3\rightarrow\mathrm{SO(3)}$ maps from the Lie algebra to the Lie group. Similarly, we redefine the "adding" operator for the camera translation $\boldsymbol{t}$ to make the perturbation $\Delta \boldsymbol{t}\in\mathbb{R}^3$ acts in the camera-local space:
\begin{equation}
    \boldsymbol{t}\oplus\Delta\boldsymbol{t}=\boldsymbol{t} + \boldsymbol{R}\Delta\boldsymbol{t},
\end{equation}
where $\boldsymbol{R}$ is the camera orientation.
For the map point positions, we express the increment in the same space of the point (\textit{i.e.}, the world).
\paragraph{Mocap-constrained Camera Tracking}
According to Eq.~\ref{eq:tracking}, without loss of generality, we respectively denote the reprojection error of point $i$ in keyframe $j$, the orientation error of keyframe $j$, and the position error of keyframe $j$ as:
\begin{equation}
\begin{array}{l}
     \boldsymbol{e}^{(\mathrm{proj})}_{ij} = \boldsymbol{x}^{\mathrm{2D}}_i - \uppi\left(\boldsymbol{R}_j^T (\boldsymbol{x}^{\mathrm{3D}}_i - \boldsymbol{t}_j)\right)\\
     \boldsymbol{e}^{(\mathrm{ori})}_{j} = \mathrm{Log}(\bar{\boldsymbol{R}}_j^T\boldsymbol{R}_j) \\
     \boldsymbol{e}^{(\mathrm{pos})}_{j} = \bar{\boldsymbol{t}}_j - \boldsymbol{t}_j.
\end{array}
\end{equation}
For brevity, we denote the map point $i$'s position in the camera $j$'s coordinate frame as $\boldsymbol{y}_{ij}=\boldsymbol{R}_j^T (\boldsymbol{x}^{\mathrm{3D}}_i - \boldsymbol{t}_j)$.
Then, the Jacobians of the errors to the pose perturbations can be computed as:
\begin{equation}\label{eq:jacobian}
\begin{array}{l}
    \left(\begin{array}{cc}
    \frac{\partial\boldsymbol{e}^{(\mathrm{proj})}_{ij}}{\partial\Delta\boldsymbol{\theta}_j} & 
    \frac{\partial\boldsymbol{e}^{(\mathrm{proj})}_{ij}}{\partial\Delta\boldsymbol{t}_j}
    \end{array}\right) = 
    \left(\begin{array}{cc}
    -\boldsymbol{J}_\mathrm{p}(\boldsymbol{y}_{ij})\boldsymbol{y}_{ij}^{\,\,\,\,\wedge} &
    \boldsymbol{J}_\mathrm{p}(\boldsymbol{y}_{ij})
    \end{array}\right)
    \\
     \left(\begin{array}{cc}
    \frac{\partial\boldsymbol{e}^{(\mathrm{ori})}_{j}}{\partial\Delta\boldsymbol{\theta}_j} & 
    \frac{\partial\boldsymbol{e}^{(\mathrm{ori})}_{j}}{\partial\Delta\boldsymbol{t}_j}
    \end{array}\right) = 
    \left(\begin{array}{cc}
    \boldsymbol{J}_\mathrm{r}\left( \mathrm{Log}(\bar{\boldsymbol{R}}_j^T\boldsymbol{R}_j)\right)^{-1} &
    \boldsymbol{O}
    \end{array}\right)
     \\
     \left(\begin{array}{cc}
    \frac{\partial\boldsymbol{e}^{(\mathrm{pos})}_{j}}{\partial\Delta\boldsymbol{\theta}_j} & 
    \frac{\partial\boldsymbol{e}^{(\mathrm{pos})}_{j}}{\partial\Delta\boldsymbol{t}_j}
    \end{array}\right) = 
    \left(\begin{array}{cc}
    \boldsymbol{O} & -\boldsymbol{R}_j
    \end{array}\right),
\end{array}
\end{equation}
where $\cdot^\wedge$ converts the 3D vector to the skew symmetric matrix, $\boldsymbol{J}_\mathrm{p}(\cdot)$ is calculated as:
\begin{equation}
\boldsymbol{J}_\mathrm{p}(\boldsymbol{y}) = \left(\begin{array}{ccc}
         \frac{f_x}{y_z} & 0 & -\frac{f_x y_x}{y_z^2}  \\
         0 & \frac{f_y}{y_z} & -\frac{f_y y_y}{y_z^2} 
    \end{array}\right),
\end{equation}
where we assume $\boldsymbol{y}=(y_x,y_y,y_z)^T$ and omit its subscript $i$ and $j$, and $f_x, f_y$ are the camera focal lengths.
$\boldsymbol{J}_\mathrm{r}(\cdot)^{-1}$ in Eq.~\ref{eq:jacobian} is calculated as:
\begin{equation}
    \boldsymbol{J}_\mathrm{r}(\theta\boldsymbol{a})^{-1}=\frac{\theta}{2}\cot \frac{\theta}{2} \boldsymbol{I}+ (1-\frac{\theta}{2}\cot \frac{\theta}{2})\boldsymbol{a}\boldsymbol{a}^T+\frac{\theta}{2}\boldsymbol{a}^\wedge,
\end{equation}
where we assume $\mathrm{Log}(\bar{\boldsymbol{R}}_j^T\boldsymbol{R}_j)=\theta\boldsymbol{a}$ in Eq.~\ref{eq:jacobian}, where $\theta\in\mathbb{R},\boldsymbol{a}\in\mathbb{R}^3$ corresponds to the angle and axis respectively.
Note that we omit the subscript $j$ for brevity.
\paragraph{Mocap-aware Bundle Adjustment}
The mocap-aware bundle adjustment (Eq.~\ref{eq:ba}) shares the same reprojection error with the camera tracking.
The absolute orientation error and the relative position error in the BA are defined as:
\begin{equation}
\begin{array}{l}
    \boldsymbol{e}^{(\mathrm{aori})}_{j} = \mathrm{Log}(\tilde{\boldsymbol{R}}_j^T\boldsymbol{R}_j) \\
    \boldsymbol{e}^{(\mathrm{rpos})}_{j}=(\tilde{\boldsymbol{t}}_j-\tilde{\boldsymbol{t}}_{\mathrm{prev}(j)}) - (\boldsymbol{t}_j-\boldsymbol{t}_{\mathrm{prev}(j)}).
\end{array}
\end{equation}
The non-zero jacobians of the two errors can be computed as:
\begin{equation}
\begin{array}{l}
    \frac{\partial\boldsymbol{e}^{(\mathrm{aori})}_{j}}{\partial\Delta\boldsymbol{\theta}_j}= 
    \boldsymbol{J}_\mathrm{r}\left( \mathrm{Log}(\tilde{\boldsymbol{R}}_j^T\boldsymbol{R}_j)\right)^{-1}\\
    \frac{\partial\boldsymbol{e}^{(\mathrm{rpos})}_{j}}{\partial\Delta\boldsymbol{t}_j} = -\boldsymbol{R}_j \\
    \frac{\partial\boldsymbol{e}^{(\mathrm{rpos})}_{j}}{\partial\Delta\boldsymbol{t}_{\mathrm{prev}(j)}} = \boldsymbol{R}_{\mathrm{prev}(j)}.
\end{array}
\end{equation}
As the map point positions are also optimized in the BA (through reprojection errors), the Jacobian of the reprojection error to the map point positions can be computed as:
\begin{equation}
    \frac{\partial\boldsymbol{e}^{(\mathrm{proj})}_{ij}}{\partial \boldsymbol{x}_i^{\mathrm{3D}}}=-\boldsymbol{J}_\mathrm{p}(\boldsymbol{y}_{ij})\boldsymbol{R}_j^T.
\end{equation}

\section{Local Map Construction for real-time tracking}\label{app:local-map}
As the reconstructed map can grow very large during tracking and finding matches between the image keypoints and 3D map points can be very time-consuming, we only use a \textit{local} visible map in the camera tracking module to ensure its real-time performance. 
Following~\cite{ORBSLAM3}, we construct the local map from the observed points in recent and neighboring frames.
One difference is that, when the visual tracking is lost, we additionally add to the local map the 3D points observed by any of \textit{1)} the candidate keyframes selected for relocalization and \textit{2)} the covisible\footnote{If two keyframes observe some shared map points, we say they are \textit{covisible}.} keyframes of the candidates.
This effectively accelerates the recovery from visual tracking losses because the camera tracking can often succeed ahead of the strict place-recognition-based relocalization since we still have an accurate camera pose due to the combination with mocap.
Therefore, we do not give up the camera tracking during visual losses as the original system does. Instead, we actively construct the local map from similar keyframes and try tracking the camera with respect to it.

\begin{algorithm}[t]
\KwIn{Global acceleration $\boldsymbol{a}_{k-1}$, camera-root position difference $\boldsymbol{p}_\mathrm{root\rightarrow cam}$ (computed from the human pose), camera localization $\boldsymbol{p}_\mathrm{cam}$ and its confidence $\boldsymbol{\varSigma}_\mathrm{cam}$ (only if the camera image is available in the current frame).}
\KwOut{The human's global position $p_k$ for the current frame.}
\BlankLine
\tcp{initialization}
\If{first frame}{
    $\hat{\boldsymbol{x}}_{k-1}\gets\boldsymbol{0}$;
    $\boldsymbol{P}_{k-1}\gets\boldsymbol{O}$\;
    initialize $\boldsymbol{A}$, $\boldsymbol{B}$, $\boldsymbol{H}$, and $\boldsymbol{Q}$ according to Eq.~\ref{eq:kfinit}\;
}
\BlankLine
\BlankLine
\tcp{prediction}
$\boldsymbol{u}_{k-1}\gets\boldsymbol{a}_{k-1}$\; 
$\hat{\boldsymbol{x}}_k^-\gets\boldsymbol{A}\hat{\boldsymbol{x}}_{k-1}+\boldsymbol{B}\boldsymbol{u}_{k-1}$\;
$\boldsymbol{P}_k^-\gets\boldsymbol{A}\boldsymbol{P}_{k-1}\boldsymbol{A}^T+\boldsymbol{Q}$\;
\BlankLine
\BlankLine
\tcp{correction}
\eIf{camera not available in the current frame}{
    $\hat{\boldsymbol{x}}_k\gets\hat{\boldsymbol{x}}_k^-$;
    $\boldsymbol{P}_k\gets\boldsymbol{P}_k^-$\;
}{
    $\boldsymbol{R}_k\gets\boldsymbol{\varSigma}_\mathrm{cam}$;
    $\boldsymbol{z}_k\gets\boldsymbol{p}_\mathrm{cam}-\boldsymbol{p}_\mathrm{root\rightarrow cam}$\;
    \eIf{relocalization detected}{
        $\hat{\boldsymbol{x}}_k\gets\hat{\boldsymbol{x}}_k^-$;
        $\boldsymbol{P}_k\gets\boldsymbol{P}_k^-$\;
        replace the position in $\hat{\boldsymbol{x}}_k$ with $\boldsymbol{z}_k$\;
    }{
       $\boldsymbol{K}_k\gets\boldsymbol{P}_k^-\boldsymbol{H}^T(\boldsymbol{H}\boldsymbol{P_k^-}\boldsymbol{H}^T+\boldsymbol{R}_k)^{-1}$\;
       $\hat{\boldsymbol{x}}_k\gets\hat{\boldsymbol{x}}_k^-+\boldsymbol{K}_k(\boldsymbol{z}_k-\boldsymbol{H}\hat{\boldsymbol{x}}_k^-)$\;
       $\boldsymbol{P}_k\gets(\boldsymbol{I}-\boldsymbol{K}_k\boldsymbol{H})\boldsymbol{P}_k^-$\;
    }
}
\BlankLine
\tcp{output}
$\boldsymbol{p}_k\gets\boldsymbol{H}\hat{\boldsymbol{x}}_k$\;
\caption{Body Translation Updater}
\label{alg:kf}
\end{algorithm}

\section{Details of Body Translation Updater}\label{app:kalman-filter}
The task of the translation updater is to estimate the drift-free human global position from the global acceleration $\boldsymbol{a}$, camera localization $\boldsymbol{p}_\mathrm{cam}$, and the camera confidence $\boldsymbol{\varSigma}_\mathrm{cam}$.
We formulate the prediction-correction algorithm in the Kalman filter framework, where the state and observation equations are defined as: 
\begin{equation}
\begin{array}{c}
     \boldsymbol{x}_k = \boldsymbol{A}\boldsymbol{x}_{k-1} +\boldsymbol{B}\boldsymbol{u}_{k-1}+\boldsymbol{\varphi}_{k-1}\\
     \boldsymbol{z}_k = \boldsymbol{H}\boldsymbol{x}_k+\boldsymbol{\psi}_k,
\end{array}
\end{equation}
where, in our setting, the state $\boldsymbol{x}=[\boldsymbol{p}^T\,\,\,\,\boldsymbol{v}^T]^T$ is the concatenation of the human's global position and velocity, $\boldsymbol{u}_{k-1}=\boldsymbol{a}_{k-1}$ is the acceleration input, $\boldsymbol{z}_k$ is the root position observation calculated from the camera localization and the estimated human pose, $\boldsymbol{\varphi}_{k-1}\sim\mathcal{N}(\boldsymbol{0}, \boldsymbol{Q})$ and $\boldsymbol{\psi}_{k}\sim\mathcal{N}(\boldsymbol{0}, \boldsymbol{R})$ denote the process and measurement noise respectively.
Then, we have:
\begin{equation}\label{eq:kfinit}
\begin{array}{c}
   \boldsymbol{A} = \left(\begin{array}{cc}\boldsymbol{I}&\Delta t \boldsymbol{I}\\ \boldsymbol{O}&\boldsymbol{I}\\\end{array}\right),\,\,\,\,\,\,\,\,
   \boldsymbol{B} = \left(\begin{array}{c}\boldsymbol{O}\\ \Delta t \boldsymbol{I}\\ \end{array}\right),\,\,\,\,\,\,\,\,
   \boldsymbol{H} = \left(\begin{array}{cc}\boldsymbol{I}&\boldsymbol{O}\\ \end{array}\right),\\
   \boldsymbol{Q} = \left(\begin{array}{cc}\boldsymbol{O}&\\&\sigma^2\boldsymbol{I}\\\end{array}\right),\,\,\,\,\,\,\,\,
   \boldsymbol{R} = \boldsymbol{\varSigma}_\mathrm{cam}.
\end{array}
\end{equation}
Readers can refer to Sec.~\ref{sec:method-translation-updater} as this equation is the matrix version of Eq.~\ref{eq:kf} and~\ref{eq:observation}.
\hl{Notice that we overload the notation $\boldsymbol{\varphi}$ here to model the full-state transition noise.}
The body translation updater algorithm is presented in Alg.~\ref{alg:kf}.
We run the updater at 60 FPS.
As the camera tracking module runs at the frame rate of the monocular camera (\textit{i.e.}, 30 FPS), we distinguish the frames \textit{with}/\textit{without} the camera image in the algorithm.
The drift-free human global positions can finally be outputted at 60 FPS.

\end{document}